\documentclass[USenglish,onecolumn,table,dvipsnames]{article}

\usepackage[utf8]{inputenc}%
\usepackage[big]{dgruyter}

\usepackage{amsmath}
\usepackage{amssymb,graphicx}

\usepackage{booktabs}
\usepackage{subcaption}
\usepackage{algorithm}
\usepackage[noend]{algpseudocode}

\usepackage{multirow}
\usepackage{adjustbox}
\usepackage{array}
\usepackage{bbm}
\usepackage{bm}

\usepackage{highlight}

\usepackage{makecell}

\usepackage{systeme,mathtools}
\usepackage{tikz}
\usetikzlibrary{shapes,decorations,arrows,calc,arrows.meta,fit,positioning}
\tikzset{
    -Latex,auto,node distance =1 cm and 1 cm,semithick,
    state/.style ={circle, draw, minimum width = 1.0 cm,inner sep=0pt},
    point/.style = {circle, draw, inner sep=0.04cm,fill,node contents={}},
    bidirected/.style={Latex-Latex,dashed},
    el/.style = {inner sep=2pt, align=left, sloped}
}

\newcommand*{\missingreference}{{\Huge \colorbox{red}{?reference?}}}
\newcommand*{\missingcitation}{{\Huge \colorbox{red}{?citation?}}}
\makeatletter
\def\@setref#1#2#3{%
  \ifx#1\relax
    \protect\G@refundefinedtrue
    \nfss@text{\reset@font\missingreference}%
    \@latex@warning{Reference `#3' on page \thepage \space
      undefined}%
  \else
    \expandafter#2#1\null
  \fi}
\def\@citex[#1]#2{\leavevmode
  \let\@citea\@empty
  \@cite{\@for\@citeb:=#2\do
    {\@citea\def\@citea{,\penalty\@m\ }%
      \edef\@citeb{\expandafter\@firstofone\@citeb\@empty}%
      \if@filesw\immediate\write\@auxout{\string\citation{\@citeb}}\fi
      \@ifundefined{b@\@citeb}{\hbox{\reset@font\missingcitation}%
        \G@refundefinedtrue
        \@latex@warning
        {Citation `\@citeb' on page \thepage \space undefined}}%
      {\@cite@ofmt{\csname b@\@citeb\endcsname}}}}{#1}}
\makeatother

\newtheorem{definition}{Definition}
\newtheorem{problem}{Problem}
\newtheorem{assumption}{Assumption}

\DeclareMathOperator*{\minl}{min\,\,}

\newcommand{\features}{\mathbf{X}}

\newcommand{\vect}[1]{\boldsymbol{\mathbf{#1}}}

\newcommand{\esttreat}{\hat{\tau}_1}
\newcommand{\estcontrol}{\hat{\tau}_2}

\newcommand{\newtext}[1]{#1}

\newcolumntype{P}[1]{>{\arraybackslash}p{#1}}
\newcolumntype{M}[1]{>{\centering\arraybackslash}m{#1}}
\newcolumntype{N}{@{}m{0pt}@{}}
\newcommand{\bftab}{\fontseries{b}\selectfont}

\def\true{true}
\def\arxiv{true}

\ifx\arxiv\true{}
\newcommand{\yesappendix}[1]{}
\newcommand{\noappendix}[1]{#1}
\else
\newcommand{\yesappendix}[1]{#1}
\newcommand{\noappendix}[1]{} 
\fi

\begin{document}

\articletype{Research Article{\hfill}Open Access}

\title{\huge Data-Driven Estimation of Heterogeneous Treatment Effects}
\runningtitle{Data-Driven HTEs}

\author*[1]{Christopher Tran}
\author[2]{Keith Burghardt}
\author[3]{Kristina Lerman}
\author[4]{Elena Zheleva}

\affil[1]{U.S. Securities and Exchange Commission; E-mail: chris.l.tran.2016@gmail.com}
\affil[2]{University of Southern California; E-mail: keithab@isi.edu}
\affil[3]{University of Southern California; E-mail: lerman@isi.edu}
\affil[4]{University of Illinois Chicago; E-mail: ezheleva@uic.edu}

\begin{abstract}
  {Estimating how a treatment affects different individuals, known as heterogeneous treatment effect estimation, is an important problem in empirical sciences. In the last few years, there has been a considerable interest in adapting machine learning algorithms to the problem of estimating heterogeneous effects from observational and experimental data. However, these algorithms often make strong assumptions about the observed features in the data and ignore the structure of the underlying causal model, which can lead to biased estimation. At the same time, the underlying causal mechanism is rarely known in real-world datasets, making it hard to take it into consideration. In this work, we provide a survey of state-of-the-art data-driven methods for heterogeneous treatment effect estimation using machine learning, broadly categorizing them as methods that focus on counterfactual prediction and methods that directly estimate the causal effect. We also provide an overview of a third category of methods which rely on structural causal models and learn the model structure from data. Our empirical evaluation under various underlying structural model mechanisms shows the advantages and deficiencies of existing estimators and of the metrics for measuring their performance.}
\end{abstract}

\journalname{Journal of Causal Inference}
\DOI{DOI}
\startpage{1}
\received{..}
\revised{..}
\accepted{..}

\journalyear{2019}
\journalvolume{1}

\maketitle

\section{Introduction}\label{sec:intro}

Estimating the effect of a treatment on an outcome is a fundamental problem in many fields such as medicine~\citep{shalit-icml17,laber-bio15,lakkaraju-aistats17}, public policy~\citep{grimmer-pa17} and more~\citep{li-www10,ascarza-jmr18}.
For example, doctors might be interested in how a treatment, such as a drug, affects the recovery of patients~\citep{foster-statmed11}, economists may be interested in how a job training program affects employment prospectives~\citep{lalonde-aer86}, and advertisers may want to model the average effect an advertisement has on sales~\cite{lewis-retail08}.

However, individuals may react differently to the treatment of interest, and knowing only the average treatment effect in the population is insufficient.
For example, a drug may have adverse effects on some individuals but not others~\citep{shalit-icml17}, or a person's education and background may affect how much they benefit from job training~\citep{lalonde-aer86,pearl-smr17}.
Measuring the extent to which different individuals react differently to treatment is known as \emph{heterogeneous treatment effect (HTE) estimation}.

Traditionally, HTE estimation has been done through subgroup analysis~\citep{gail-bio95,bonetti-bio04}.
However, this can lead to cherry-picking since the practitioner is the one who identifies subgroups for estimating effects.
Recently, %
there has been more focus on \emph{data-driven} estimation of heterogeneous treatment effects by letting the data identify which features are important for treatment effect estimation  using machine learning techniques~\citep{johansson-icml16,louizos-neurips17,shalit-icml17,tran-aaai19}.
A straightforward approach is to create interaction terms between all covariates and use them in a regression~\citep{athey-handbook17}.
However, this approach is limited to low-dimensional datasets and is not practical for large datasets with many features.
To deal with the high dimensionality of data, many supervised learning-based methods have been developed and adapted to the problem of HTE estimation~\citep{su-jmlr09,hill-jcgs11,imai-annals13,athey-stat15,johansson-icml16,athey-pnas16,shalit-icml17,tran-aaai19,kunzel-pnas19}.

HTE estimation methods can be broadly categorized into counterfactual prediction and effect estimation methods.
A counterfactual prediction model learns to predict each unit's potential outcomes when treated and not treated. 
In turn, those predictions can be used to estimate a treatment effect~\citep{foster-statmed11,shalit-icml17,kunzel-pnas19}.
A heterogeneous effect estimation method learns a model that predicts the treatment effect directly~\citep{su-jmlr09,athey-pnas16,tran-aaai19}.
Effect estimation models can be learned through specialized objective functions~\citep{su-jmlr09,athey-pnas16} or in tandem with counterfactual prediction~\citep{louizos-neurips17,kunzel-pnas19}.
When data is collected from randomized controlled trials where pre-treatment variables are clearly defined, including these variables in HTE estimation can be effective and lead to valid estimations.
However, data from controlled trials is not always available, and some studies have focused on estimating heterogeneous treatment effects from observational data~\citep{shalit-icml17,athey-annals19,xie-soc12,lada2019observational,veitch2020adapting,syrgkanis2021causal,econml}. In such studies, it is not always known which variables are pre-treatment variables and HTE estimation methods may use all variables in the data, relying on the ignorability assumption~\citep{montgomery-ajps18}.
In Section~\ref{sec:structural_causal_models} and in our experiments, we discuss when including all variables can lead to errors in HTE estimation.

A third category of methods, which rely on structural causal models (SCMs)~\citep{pearl-book09}, provide criteria for the selection of variables that allow for unbiased causal effects. SCMs are graphical models that model explicitly the cause-effect relationships between variables.
The selection of variables through SCMs can help data-driven HTE estimation methods better estimate causal effects, as we show in our experiments.
However, the underlying causal mechanism of data and its SCM are rarely known in real-world settings. To circumvent this problem, one can use a causal structure learning algorithm~\citep{spirtes-book00} to learn a skeleton of the underlying causal model from the data itself.
This survey provides an overview of recently developed data-driven HTE estimation methods, and an overview of SCMs and how they can be used for HTE estimation.

The rest of the paper is organized as follows. 
Section~\ref{sec:background} gives basic background knowledge.
Section~\ref{sec:survey} reviews counterfactual prediction and effect estimation methods. %
Section~\ref{sec:structural_causal_models} provides an overview of data-driven HTE estimation with structural causal models.  %
Section~\ref{sec:experiments} empirically evaluates HTE estimators on several types of datasets.
Section~\ref{sec:conclusion} concludes the paper and discusses open problems.

\section{Heterogeneous Treatment Effects}\label{sec:background}

In this section, we provide background on heterogeneous treatment effect estimation. We define the problem of heterogeneous treatment effect estimation and common assumptions to allow for estimation.
We introduce structural causal models~\citep{pearl-book09} and how they can be used to assess heterogeneity.

\subsection{Data model}\label{sec:data_model}

Before formalizing the problem and data model, we introduce the notation used in this survey.
We denote a space of random variables with a calligraphic letter (e.g., $ \mathcal{X} $).
We denote random variables with a capital letter (e.g., $Y$) and random vectors with a boldfaced capital letter (e.g., $\mathbf{X}$).
Subscripts signify the random variable associated with a specific unit (e.g., $Y_i$ or $\mathbf{X}_i$).
Lowercase letters denote an assignment of a random variable or vector (e.g., $Y_i = y$ or $\mathbf{X}_i = \mathbf{x}$).
Each unit $i$ has a vector of ``pre-treatment'' features (or covariates), $\mathbb{X}_i \in \mathcal{X}$, a binary treatment indicator, $T_i$, where $T_i=1$ indicates that $i$ is treated and $T_i=0$ that is not treated, and two potential outcomes, one when treated, $Y_i(T_i=1)$, and one when not treated, $Y_i(T_i=0)$. 
For brevity, we write $Y_i(T_i=0)$ and $Y_i(T_i=1)$ as $Y_i(0)$ and $Y_i(1)$, respectively.
Most work on heterogeneous treatment effect estimation considers binary treatment and for simplicity we will consider binary treatment throughout this survey as well. Recent work has also considered how to estimate heterogeneous effects when the treatment is non-binary~\citep{tran-aaai19,schwab-arxiv19,kennedy-jrss17}
The underlying data distribution is represented by a tuple $(\features_i, T_i, Y_i(0), Y_i(1))^N$ of $N$ independent and identically distributed individuals.
The fundamental problem of causal inference is that for each individual, we can observe either the potential outcomes under treatment ($Y_i(1)$) or under control ($Y_i(0)$), but not both.
Therefore, the observed dataset as $\mathcal{D} = (\mathbf{x}_i, t_i, y_i)^N$ with $N$ individuals where $Y_i$ is defined as:
\begin{equation}
    \label{eq:potential_outcomes}
    y_i = t_i Y_i(1) + (1 - t_i) Y_i(0).
\end{equation}
We denote $ Y_i $ as the factual outcome and the counterfactual outcome, $ Y_i' $, as the outcome that would have occurred if the treatment was different.
We define the treatment group as all individual units $i$ such that $t_i = 1$ and the control group as all units $i$ such that $t_i = 0$.

\subsection{Heterogeneous Treatment Effect Estimation}\label{sec:hte}

The \emph{individual treatment effect} (ITE) for unit $i$ is defined as the difference in the unit's potential outcomes, denoted by $\tau_i$:
\begin{equation}
    \label{eq:ite}
    \tau_i := Y_i(1) - Y_i(0).
\end{equation}
Due to the fundamental problem of causal inference, $\tau_i$ cannot be observed for any individual unit and several estimands have been developed to help with its estimation. The first estimand focuses on estimating the average treatment effect (ATE) over the population and it is defined as the expected difference in potential outcomes:
\begin{equation}
    \label{eq:ate}
    \text{ATE} := E[Y(1) - Y(0)].
\end{equation}

While the average treatment effect would give us a good idea about the average effect of the units, it does not always provide a good estimate of the ITE for any one individual unit. When different units react differently to treatment and have different ITEs, i.e., \emph{heterogeneous treatment effects} (HTEs), a more appropriate estimand is the conditional average treatment effect (CATE):
\begin{equation}\label{eq:cate}
    \tau(\mathbf{x}) := \mathbb{E} \left[ Y(1) - Y(0) \mid \features = \mathbf{x} \right].
\end{equation}
HTE estimation assumes that similar units have similar treatment effects, and CATE captures the expected treatment effect for units that have the same features $\mathbf{x}$.
An HTE estimator, defined as $\hat{{\tau}}(\mathbf{X})$, seeks to estimate the CATE defined in equation~\eqref{eq:cate}.
As shown by~\citep{kunzel-pnas19}, the best estimator for CATE is also the best estimator for ITE in terms of the mean squared error. However, it is important to note that the fact that CATE captures the expected effect value for units with certain characteristics does not imply that the ITE of any given unit with these characteristics is equal to its estimated CATE. Since the ITE can never be identified, we focus on CATE estimation for the rest of the paper. 

The goal of HTE estimation is to develop estimators of CATE which have the smallest root mean squared error for the ITE\@.
\begin{problem}\label{prob:hte_problem}
(Heterogeneous treatment effect estimation)
Given a dataset $\mathcal{D}=(\features_i, Y_i, T_i)$ of N instances with some true ITE, $\vect{\tau} = (\tau_i)^N$, estimate CATE $\hat{\tau}(\mathbf{X})$ such that the root-mean-squared error (RMSE) is minimized:
\begin{equation}\label{eq:cate_objective}
    \minl_{\hat{\tau}} \text{RMSE} = \frac{1}{N} \sum_{i=1}^N  (\tau_i - \hat{\tau}(\mathbf{X}_i)) ^ 2.
\end{equation}
\end{problem}
In practice, equation~\eqref{eq:cate_objective} cannot be estimated in real-world data because the ground truth for $tau_i$ is unknown.
Thus, HTE estimators are typically evaluated on synthetic or semi-synthetic datasets~\citep{curth-neurips21} where the outcomes are carefully generated and $tau_i$ is known. However, the performance on synthetic datasets may not always correspond to the performance of these estimators in real-world scenarios.
To address this concern, a line of research has focused on developing heuristic metrics that may correlate with RMSE and help with evaluation on real-world data, \citep{schuler-model18,saito-cv19,schwab-perfect18}. We describe them in Section~\ref{sec:data_driven_background}.

\subsection{HTE Assumptions}\label{sec:hte_assumptions}

To estimate HTEs from observational data, some assumptions on the set of features and data given are needed.
We present some common assumptions used in heterogeneous treatment effect estimation.

\begin{assumption}\label{assumption:strong_ignorability}
    (Ignorability) Given a binary treatment $T$, potential outcomes $Y(0)$, $Y(1)$, and a vector of features $\features$, ignorability states:
    \begin{align}
        \{ Y(0), Y(1) \} \perp T \mid \features,
    \end{align}
\end{assumption}
This states the \emph{potential} outcomes are conditionally independent of the treatment conditioned on the features.
When ignorability holds, $\features$ is said to be ``admissible'', meaning treatment effects can be estimated without bias.
This assumption is also known as unconfoundedness or the assumption of no hidden confounding.

\begin{assumption}\label{assumption:overlap}
    (Overlap) For every unit $i$, $0 < P(T_i = 1) < 1$ with probability 1.
\end{assumption}
This assumption states that all units have a non-zero probability of assignment to each treatment condition.
Assumption~\ref{assumption:overlap} is also known as positivity or common support.

Another common assumption used is the Stable Unit Treatment Value Assumption (SUTVA) or the assumption of no interference~\citep{cox-book58,rubin-jasa80}.
\begin{assumption}\label{assumption:sutva}
    (Stable Unit Treatment Value Assumption (SUTVA)) Given any two units $i$ and $j$, the potential outcomes of $i$, $\{Y_i(0), Y_i(1) \}$ are unaffected by the treatment assignment of unit $j$:
    \begin{equation}
        \{Y_i(0), Y_i(1)\} \perp T_j, \forall j \neq i.
    \end{equation}
\end{assumption}
With the SUTVA assumption, individual outcomes are not affected by the treatment or outcomes of peers.
Though there has been some work focused on estimating HTEs without the SUTVA assumption~\citep{yuan-www21,tran-aaai22}, a majority of work in HTE estimation has focused on non-networked settings, which we focus on in this survey.

\section{Counterfactual prediction and effect estimation}\label{sec:survey}

This section provides a survey of existing heterogeneous treatment effect estimation methods.
We categorize the methods into two main categories: methods that perform counterfactual prediction and methods that estimate treatment effect directly.

Counterfactual prediction methods predict counterfactual outcomes when an individual is treated and not treated~\citep{johansson-icml16,shalit-icml17,kunzel-pnas19}.
Then an estimate of an individual treatment effect is the difference in predicted outcomes.
The second category consists of methods that perform effect estimation directly.
These methods use counterfactual predictions as part of a process to learn an effect estimator~\citep{athey-pnas16,tran-aaai19,athey-annals19} or use different objective functions to estimate effects~\citep{kunzel-pnas19,louizos-neurips17}.
Table~\ref{tab:method_overview} shows subcategories of each category, together with representative papers.

\begin{table}[h]
    \centering
    \caption{An overview of HTE estimator categories and subcategories.}\label{tab:method_overview}
    \begin{tabular}[t]{M{0.47\linewidth}  M{0.47\linewidth}}
        Counterfactual Prediction & Effect Estimation \\
        \midrule
        Single-Model Approaches \citep{hill-jcgs11,kunzel-pnas19,athey-stat15, atan-aaai18} & Transformed Outcome-based Approaches \citep{rosenbaum-bio83,athey-stat15} \\
        {Two-Model Approaches} \citep{foster-statmed11,cai-bio11,athey-stat15,kunzel-pnas19, johansson-icml16,shalit-icml17}  & {Meta-Learning Approaches} \citep{xie-soc12,powers-statmed18,kunzel-pnas19,kennedy-arxiv20} \\
        {Multi-Model Approaches} \citep{schwab-arxiv19,shi-neurips19} & {Tree-based Approaches} \citep{athey-pnas16,tran-aaai19,su-jmlr09} \\
        & {Forest-based Approaches} \citep{wager-jasa18,athey-annals19,powers-statmed18} \\
        & {Deep Learning Approaches} \citep{yoon-iclr18,louizos-neurips17} \\
    \end{tabular}
\end{table}

\subsection{Counterfactual prediction}\label{sec:counterfactuals}

The first category of methods predicts counterfactuals from data, the outcome under treatment and under control, $ \hat{Y}(1), \hat{Y}(0) $, respectively, and uses these predictions for effect estimation, $ \hat{Y}(1)-\hat{Y}(0)$.

\subsubsection{Single-model approaches}

The simplest way to make counterfactual predictions is by using a \emph{single-model approach}.
The single-model approach estimates CATE using a single regression or classification model.
Given the data \( D = (\mathbf{X}_i, Y_i, T_i) \), a supervised method is trained to predict \( Y_i \) given \( \mathbf{X}_i, T_i \), \( \hat{\mu}(T_i, \mathbf{X}_i) = \hat{Y}_i \).
The CATE can be predicted as the difference between the prediction with \( T_i = 1 \) and \( T_i = 0 \).
\begin{equation}
    \hat{\tau}(\mathbf{x}) = \hat{\mu}( T = 1, \mathbf{X} = x) - \hat{\mu}(T = 0, \mathbf{X} = x).
\end{equation}

Hill proposed using Bayesian Additive Regression Trees (BART) to estimate heterogeneous treatment effects in a flexible, nonparametric way~\citep{hill-jcgs11}. 
BART models the response surface using a sum of regression trees, automatically handling nonlinearities and interactions without the need to explicitly specify them. 
The Bayesian framework provides coherent uncertainty intervals and regularization priors to avoid overfitting. 
A key advantage of BART is its simplicity of use, requiring minimal user input compared to many other nonparametric methods.
Athey and Imbens used regression trees as the single-model approach~\citep{athey-stat15}.
K{\"{u}}nzel et al.\ used a single-model approach with gradient-boosted trees and random forests~\citep{kunzel-pnas19} and name these single-model approaches S-Learner.

Since the single-model approach only requires regression or classification, any off-the-shelf supervised learning algorithm can be used.
However, the single-model approach is unable to capture nuanced differences between treatment and control groups. Because there is a single treatment response function for both the treatment and control groups and the treatment variable is treated as another feature, the learned function has the same covariate weights learned for both treatment and control units. 

Besides methods using classical machine learning methods, several deep learning approaches for estimating HTEs have been proposed recently using the single-model approach.
Atan et al.\ propose a deep learning model that consists of two stages~\citep{atan-aaai18}.
In the first stage, Deep-Treat uses a bias-removing auto-encoder to learn a balanced representation, $ \Phi(\mathbf{x}) $, of treatment and control groups.
The balancing is learned through cross-entropy loss between the marginal and conditional treatment distributions given the hidden representation.
Deep-Treat trains a neural network to predict the outcome using the learned representation and the treatment in the second state.

\subsubsection{Two-model approaches}

In contrast to the single-model approach, another way to model heterogeneity is to model two potential outcomes separately, called the \emph{two-model approach}.
In the two-model approach, a separate model is built on treated and control groups.
Then, to estimate CATE, the difference between treatment group model prediction and control group model prediction is used to estimate the effect.  In the two-model approach, there are two separate treatment response functions, each one modeling treatment or control units. This allows different covariates to have different influences on the outcome for the treatment group than the ones for the control group.

\begin{figure}[h]
    \centering
    \begin{subfigure}[b]{0.45\linewidth}
        \centering
        \includegraphics[width=\textwidth]{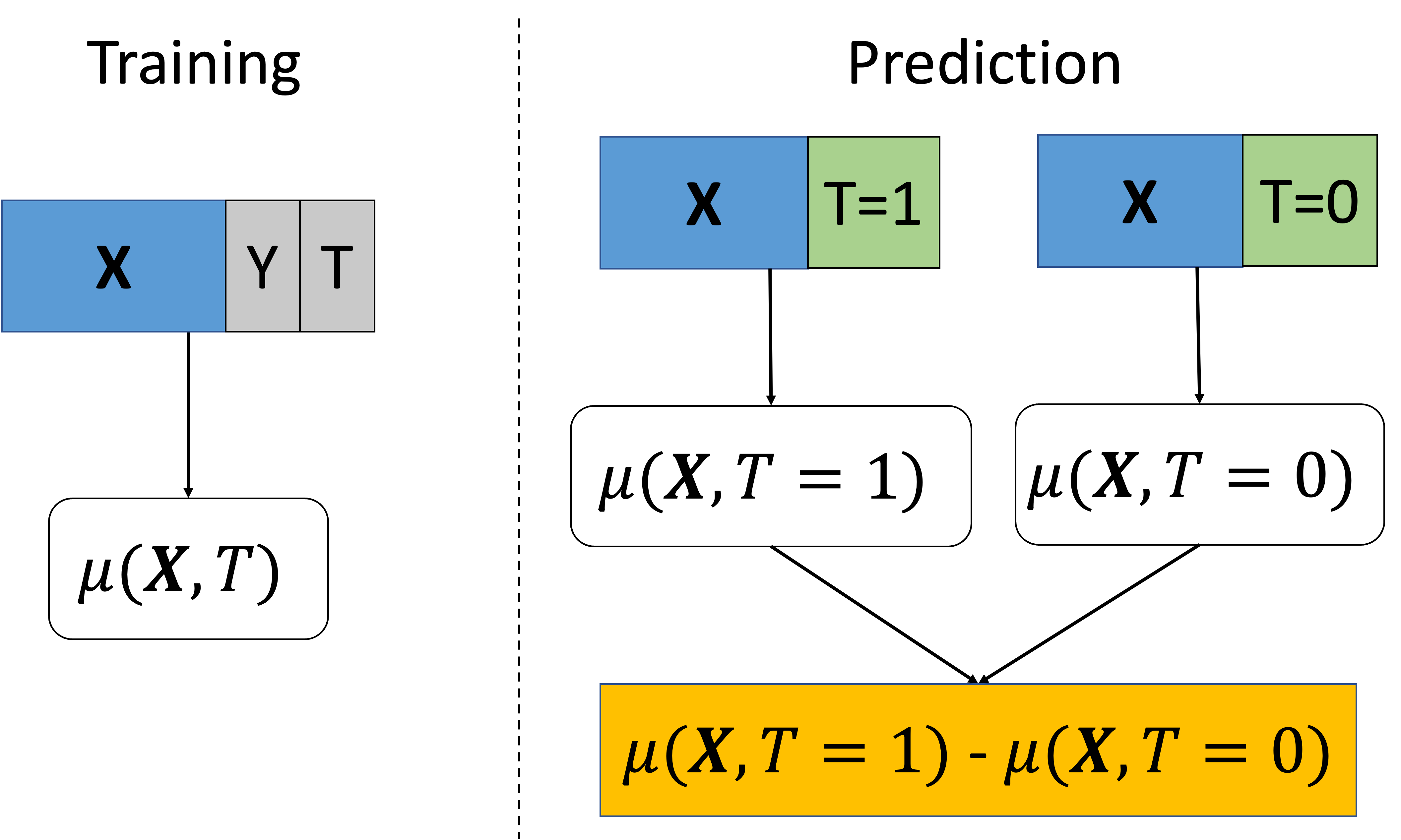}
        \caption{Single-model approach.}\label{fig:slearner}
    \end{subfigure}
    \hfill
    \begin{subfigure}[b]{0.45\linewidth}
        \centering
        \includegraphics[width=\textwidth]{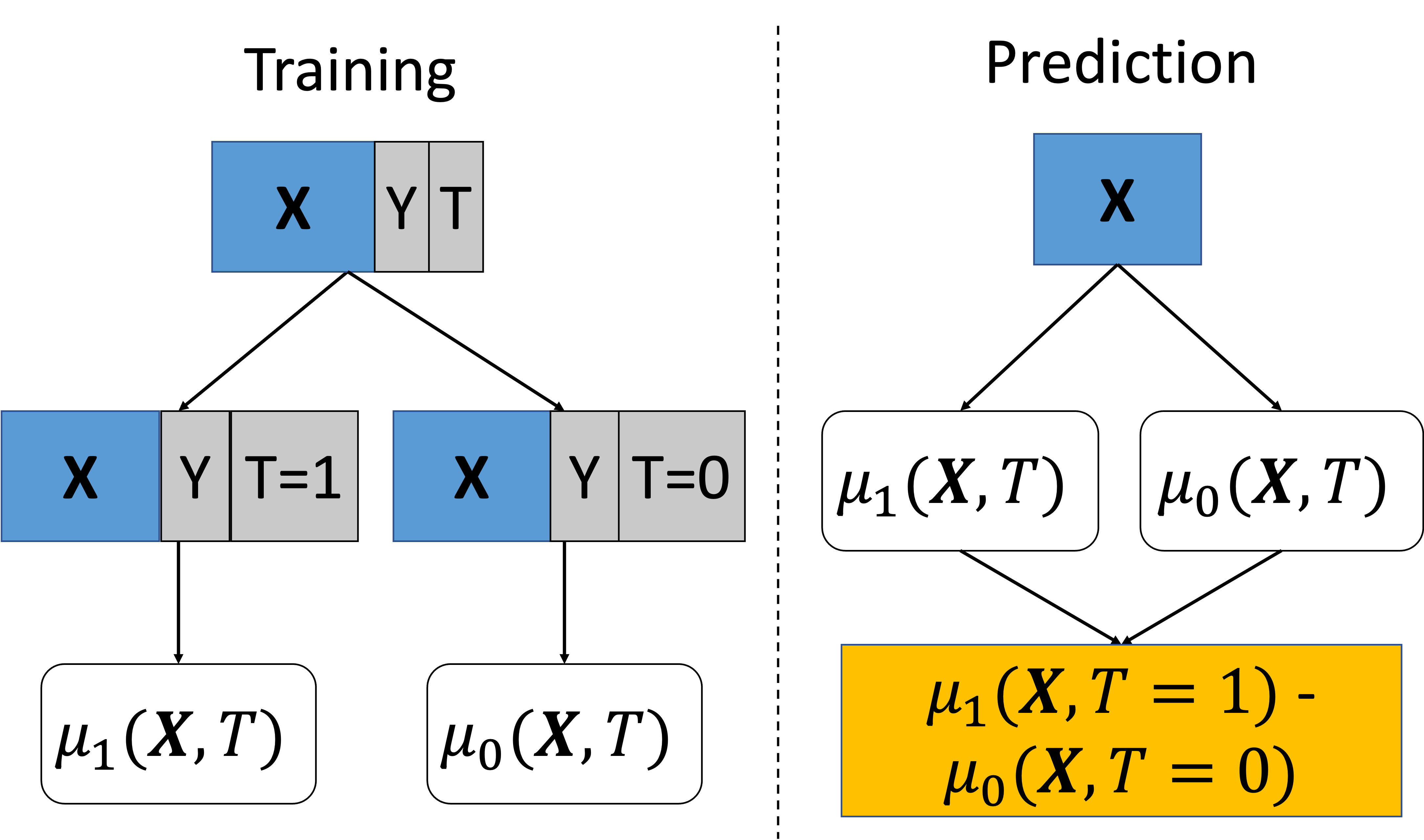}
        \caption{Two-model approach.}\label{fig:tlearner}
    \end{subfigure}
    \caption{A comparison of training and prediction of effects using single-model and two-model approaches.
    The single-model approach uses one estimator for training, and for prediction, features are appended with a treatment indicator of 1 and a treatment indicator of 0 for two separate predictions based on the single model. Then the effect is estimated as the difference.
    In the two-model approach, training is done using two separate models for treated and control groups.
    For prediction, the features are input into the two separate models, and a difference is computed as the predicted effect.}\label{fig:slearner_tlearner}
\end{figure}

Formally, let \( \esttreat(\mathbf{x}) \) and \( \estcontrol(\mathbf{x}) \) be supervised estimators for the outcome for the treated and control groups, respectively.
The CATE is estimated as:
\begin{align}
    \hat{\tau}(\mathbf{x}) & = E[Y \mid T = 1, \mathbf{X} = \mathbf{x}] - E[Y \mid T = 0, \mathbf{X} = \mathbf{x}] \\
    & = \esttreat(\mathbf{x}) - \estcontrol(\mathbf{x}).
\end{align}
The main difference between the two-model and single-model approaches is that the two-model approach trains two separate estimators for treatment and control outcome prediction instead of one model for both.
The two model approach has been used with Linear Regression~\citep{cai-bio11}, Trees~\citep{athey-stat15,hill-jcgs11}, Random Forests~\citep{foster-statmed11}, and Gradient Boosted Trees~\citep{kunzel-pnas19}.
Figure~\refeq{fig:slearner_tlearner} contrasts how single-model and two-model approaches perform training and validation.

\begin{figure}[t]
    \centering
    \includegraphics[width=0.7\textwidth]{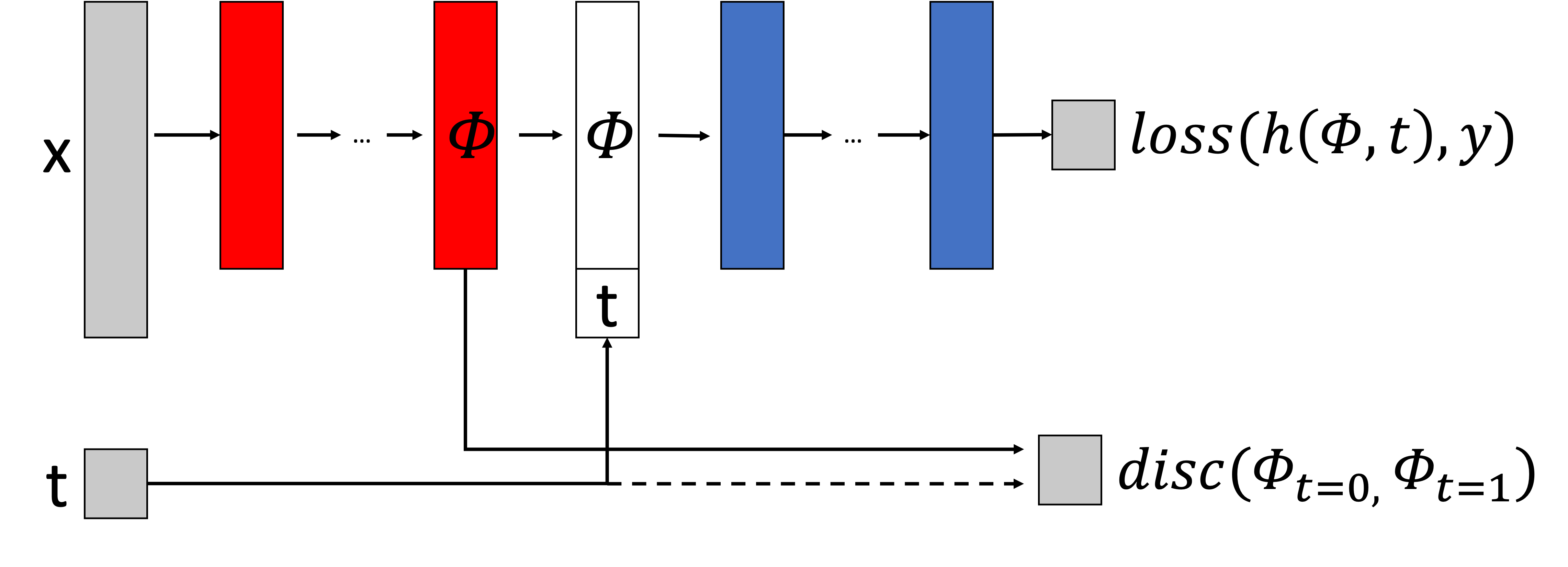}
    \caption{A high-level architecture of the Balancing Neural Network (BNN) from~\citep{johansson-icml16}. The first part learns a representation of the features only, $ \Phi $. 
    In the second part, the treatment is appended, and the neural network is optimized for prediction performance and minimizing the representations of treated and control representations.}\label{fig:bnn}
\end{figure}

In addition, recent work has applied the two-model approach with deep learning estimators.
Johansson et al.\ proposed Balancing Neural Network (BNN), which aims to leverage representation learning and balanced representations for treatment and control groups~\citep{johansson-icml16}.
They learn a neural network architecture with two connecting parts to predict the outcome.
The first part learns the representation of the features, $ \Phi(\mathbf{x}) $.
In the second part, they append the treatment to the hidden representation and learn a representation to predict the outcome.
Separately, they constrain $ \Phi(\mathbf{x}) $ to be similar between treated, $ \Phi_{t=1}(\mathbf{x}) $, and control, $ \Phi_{t=0}(\mathbf{x})$.
The intuition is that the hidden representations should be ``balanced''.
The basic BNN architecture is shown in Figure~\ref{fig:bnn}.

Shalit et al.\ extend the idea of BNNs to the Treatment-Agnostic Representation Network (TARNet) and Counterfactual Regression (CFR)~\citep{shalit-icml17}.
They extend the discrepancy metric by introducing a more flexible family of algorithms in Integral Probability Metrics (IPM) for balancing distributions.
Instead of considering a one-headed network that predicts an outcome based on appending the treatment to the hidden representation, they consider two heads predicting outcome when treated and not treated separately.
The intuition is that the treatment indicator appended to $ \Phi(\mathbf{x}) $ may lose importance in the second part of the network, and so explicitly separating the final predictions can capture differences in treatment and control.

Liu et al.\ propose Split Covariate Representation Network (SCRNet)~\citep{liu-acml20}.
In this work, they propose to divide the covariate space into four categories: confounders, instrumental variables, adjustment variables, and irrelevant variables.
Their goal is to identify the category of each feature and use only relevant features for training a neural network.
The intuition is that irrelevant features are unnecessary for estimation and instrumental variables may bias effect estimations~\citep{pearl-arxiv12,myers-aje11,middleton-pa16}.
Instead, they identify confounders, which confound or are parents of the treatment in outcome in an SCM, and parents of outcome only which are referred to as adjustment variables.
Then, they train a neural network where confounders and adjustment variables are input separately and have their own hidden representations before being concatenated and separated into a two-head network, similar TARNet.
The main difference is the splitting of the initial input into confounding and adjustment variables.

\subsubsection{Multi-Model Approaches}

Related to the TARNet approach~\citep{shalit-icml17}, Schwab et al.\ propose Perfect Match~\citep{schwab-arxiv19}.
They extend TARNet in two ways.
First, they propose using $ k $ heads to predict multiple treatment outcomes.
Second, they propose a new way to optimize the neural network model based on nearest neighbor matching and the precision in estimation of heterogeneous effect (PEHE)~\citep{hill-jcgs11}, which measures the mean-squared error in HTE estimation, called NN-PEHE\@.
They propose computing an imputed treatment effect based on nearest neighbors.
Since the true causal effect is unknown for individual units, they substitute the true counterfactual outcome with the outcome from the nearest neighbor.
Their goal is to minimize NN-PEHE for training Perfect Match.
Shi et al.\ proposed another extension of TARNet, called Dragonnet~\citep{shi-neurips19}. 
Instead of a two-head neural network, they introduce an additional head which predicts the propensity score from the hidden layer.
In addition, they introduce a modification of the objective function for training neural networks called targeted regularization.

\subsection{Effect estimation}\label{sec:effect_estimation}

In contrast to methods that perform counterfactual prediction, effect estimation methods estimate the treatment effect directly.
They include methods that utilize supervised learning~\citep{athey-stat15,kunzel-pnas19,yoon-iclr18} and methods that have tailored objectives for learning HTEs~\citep{athey-pnas16,tran-aaai19,wager-jasa18}.

\subsubsection{Transformed outcomes}

In the transformed outcomes approach, the outcome variable is transformed such that a regression or classification method can be used to predict the transformed outcome.
Any off-the-shelf estimator can be directly applied since the CATE is estimated directly from the transformed outcome.

Under the assumption of unconfoundedness, the transformed outcome is~\citep{athey-stat15}:
\begin{equation}
    Y_i^* = Y_i \cdot \frac{T_i - e(\mathbf{X}_i)}{e(\mathbf{X}_i) \cdot (1 - e(\mathbf{X}_i) )},
\end{equation}
where \( Y_i^* \) is the transformed outcome and \( e(\mathbf{x}) = \Pr(T_i = 1 \mid \mathbf{X}_i = \mathbf{x}) \) is the propensity score~\citep{rosenbaum-bio83}.
In practice, the propensity score is estimated from data using a supervised estimator such as logistic regression~\citep{setoguchi-pharm08}.
In this setting, the CATE is estimated as the estimated conditional mean of the transformed outcome.
Using the transformed outcome, the conditional mean of the predicted outcome is equal to the CATE:
\begin{equation}
    \hat{\tau}_{to}(\mathbf{x}) = \mathbb{E}[Y^{*} \mid \mathbf{X} = \mathbf{x}].
\end{equation}
Transformed outcomes has been used with regression trees~\citep{athey-stat15} and for building transformed outcome forests~\citep{powers-statmed18}.

\subsubsection{Meta-learning}

Meta-learners or meta-algorithms build upon supervised learning methods, called base learners, to estimate CATE\@.
Two special cases of meta-algorithms are the single-model and two-model approaches, as they take supervised base learners and utilize their outputs to estimate CATE\@.
Xie et al.\ propose three algorithms based on propensity scores: Stratification-Multilevel (SM), Matching-Smoothing (MS), and Smoothing-Differencing (SD)~\citep{xie-soc12}.
The common trend of these algorithms is to estimate propensity scores using a supervised learning method and create strata based on propensity scores and covariates.
In SM, stratum-specific treatment effects are estimated, and a trend across strata is obtained through variance-weighted least squares regression.
For MS, treated units are matched to control units with a matching algorithm to estimate the true effect, and a nonparametric model is applied to the matched effects.
Finally, for SD, nonparametric regression is used to fit the outcome on the propensity score.
To estimate heterogeneous effects, take the difference between the nonparametric regression line between treated and untreated at different levels of the propensity score.

Powers et al.\ propose Pollinated Transformed Outcome (PTO) forests~\citep{powers-statmed18}.
First, they use a transformed outcome approach and build a random forest.
Then, predictions in trees are replaced with the estimate of $ \hat{\mu}(T=1, \mathbf{x}) $ and $ \hat{\mu}(T=0, \mathbf{x}) $ as in the single model approach.
Finally, an additional random forest is build to predict $ \hat{\tau} $.

K{\"{u}}nzel et al.\ propose X-Learner, which aims to improve upon the two-model approach~\citep{kunzel-pnas19}.
The goal of the X-Learner is to use information from the control group to learn better estimates for the treatment group and vice versa.
X-Learner consists of three steps.
The first step is to learn two supervised estimators $ \hat{\mu}_0 $ and $ \hat{\mu}_1 $. 
This is the same as the two-model approach.
Step two involves imputing the treatment effects from predictions from the opposite group by taking the difference between factual and predicted counterfactual outcomes:
\begin{gather}
    \tilde{D}_i^1 = Y_i^{1} - \hat{\mu}\big(\mathbf{X}_i^{1} \big), \\
    \tilde{D}_i^0 = \hat{\mu}\big( \mathbf{X}_i^{0} \big) - Y_i^{0},
\end{gather}
where $ \tilde{D}_i^1 $ and $ \tilde{D}_i^0 $ is the imputed treatment effect for unit $ i $ if they were treated or untreated, respectively.
Then, two estimators $ \hat{\tau}_0(\mathbf{x}) $ and $ \hat{\tau}(\mathbf{x}) $, are trained on the imputed datasets, $ \tilde{D}_i^1 $ and $ \tilde{D}_i^0 $, respectively.
Finally, the CATE is estimated by a weighted average of the two estimators:
\begin{equation}
    \hat{\tau}(\mathbf{x}) = g(\mathbf{x}) \hat{\tau}_0(\mathbf{x}) + (1 - g(\mathbf{x})) \hat{\tau}_1(\mathbf{x}),
\end{equation}
where $ g \in [0,1]$ is a weight function.
K{\"{u}}nzel et al.\ suggest using an estimate of the propensity score for $ g $, so that $ g = \hat{e}(\mathbf{x}) $~\citep{kunzel-pnas19}.

Kennedy proposes a two-stage doubly robust method for HTE estimation called Double Robust Learner (DR-Learner)~\citep{kennedy-arxiv20}.
The training data is randomly separated into three sets: $ D_{1a}^k, D_{1b}^k, D_{2}^k $ each containing $ k $ observations.
In the first step, two estimators are trained on $ D_{1a}^k $ and $ D_{1b}^k $, called \emph{nuisance} training. 
An estimator of the propensity scores, $ \hat{e} $, is constructed on $ D_{1a} $ and estimates of $ (\hat{\mu}_0 $ and $ \hat{\mu}_1) $ are constructed by performing prediction $ D_{1b} $.
Then, a pseudo-outcome is constructed as:
\begin{equation}
    \hat{\varphi}(\mathbf{X}, Y, T) = \frac{T - \hat{e}(\mathbf{X})}{\hat{e}(\mathbf{X}) (1 - \hat{e}(\mathbf{X})) } \Big( Y - \hat{\mu}_T(\mathbf{X}) \Big) + \hat{\mu}_1(\mathbf{X}) + \hat{\mu}_0(\mathbf{X}).
\end{equation}
Finally, regressing the pseudo-outcome on $ \mathbf{X} $ in the test sample $ D_{2}^k $ obtains estimations of the CATE\@:
\begin{equation}
    \hat{\tau}_{dr}(\mathbf{x}) = \mathbb{E}[\hat{\varphi}(\mathbf{X}, Y, T) \mid \mathbf{X}=\mathbf{x}].
\end{equation}
It is recommended to cross-fit with the three sets and use the average of the resulting CATE estimators.

\subsubsection{Tree and forest-based methods}

\newtext{
Tree and forest-based methods for HTE estimation are designed to discover groups within the data sample with different causal effects. Su et al.\ developed a tree-based approach drawing on the CART algorithm, where binary covariate splits are determined by comparing mean differences through a t-test~\citep{su-jmlr09}. They introduced a key interaction measure for assessing treatment interaction within splits:
\begin{equation}
g(s) = \frac{(\bar{y}^L_1 - \bar{y}^L_0) - (\bar{y}^R_1 - \bar{y}^R_0)}{\hat{\sigma} \sqrt{1/n_1 + 1/n_2 + 1/n_3 + 1/n_4}},
\end{equation}
where \( g(s) \) relates to the splitting criterion and follows a \( \chi^2 \) distribution in large samples.
}

Athey and Imbens used causal trees to maximize observed subgroup treatment effects, highlighting heterogeneity. They adopted an adaptive criterion to optimize splits, defined by:
\begin{equation}
F_a(\ell) = N_\ell \cdot \hat{\tau}^2(\ell),
\end{equation}
where \( \hat{\tau}(\ell) \) represents the average causal effect in a node, \(\ell \). The honest approach further refines this process, splitting the data into training and estimation sets and penalizing high variances in outcomes.

Tran and Zheleva proposed using validation sets to adjust causal trees, called CT-L, creating a cost term for penalizing discrepancies between training and validation data effects:
\begin{equation}
C(\ell) = N_\ell^{val} \cdot | \hat{\tau}(\ell^{tr}) - \hat{\tau}(\ell^{val}) |,
\end{equation}
Thus, the partition measure, $F_a(\ell)$ for CT-L is adjusted by this cost term.
They also introduced the notion of triggers to find the treatment levels or thresholds that yield maximal effects for subpopulations.

Wager and Athey proposed a random forest algorithm called Causal Forest for estimating heterogeneous treatment effects~\citep{wager-jasa18}.
Causal Forest constructs a forest of honest Causal Trees that each provides a CATE estimation. Then, an average of the predicted CATE from all trees is used as the final prediction.
They showed that the estimation of Causal Forest is asymptotically Gaussian and unbiased for the true CATE if the base learner used is \emph{honest}~\citep{wager-jasa18}.
Athey et al.\ extended this idea to Generalized Random Forests~\citep{athey-annals19}.
Generalized random forests extend random forests into a method for estimating any quantity through local moment conditions and can be used for heterogeneous treatment effect estimation.

\subsubsection{Deep learning approaches}

Several other deep learning approaches have also been developed based on different architectures.
Louizos et al.\ consider a different perspective and motivation for studying hidden representations for HTE estimation~\citep{louizos-neurips17}.
Instead of considering the given features $ \mathbf{X} $ as confounders, they consider it as noisy views on the hidden confounder $ \mathbf{Z} $.
Their goal is to learn the representation of $ \mathbf{Z} $ for causal effect adjustment using the observed features $ \mathbf{X} $.
To do this, they use variational autoencoders to learn $ \mathbf{Z} $ for the treatment and control groups, and use the hidden representation for prediction of outcomes.

Yoon et al.\ propose Generative Adversarial Networks for Individualized Treatment Effects (GANITE)~\citep{yoon-iclr18}.
GANITE has two components, a counterfactual block which generates counterfactual predictions, and an ITE block that generates CATE predictions given the counterfactual prediction.
This work leverages Generative Adversarial Networks (GANs) in each block to create a better counterfactual dataset.
After training the counterfactual block, generated data is fed to the ITE block which uses a GAN to learn to generate HTEs.

A final important note to mention is that consistent estimation of the CATE requires consistent estimation of the outcome or propensity score model, depending on the model~\citep{funk-aje11,robins-jasa94}. 
For example, many of the counterfactual prediction methods discussed so far rely on the correct estimation of the predicted outcomes. 
Some work has targeted this gap through doubly-robust estimation~\citep{funk-aje11,robins-jasa94}, such as DR-Learner~\citep{kennedy-arxiv20}, or through usage of propensity scores in tandem with counterfactual predictions akin to doubly-robust estimation, such as Dragonnet~\citep{shi-neurips19}, discussed previously.
In addition, methods that use machine learning estimation should include careful consideration of the models used for outcome estimation, as some machine learning models do not provide consistent estimation, trading bias for reduced variance.

\subsection{Metrics for empirical evaluation of HTE estimators}\label{sec:data_driven_background}

Since the true ITE is unknown in real-world scenarios, several heuristic HTE evaluation metrics have been developed for evaluating the performance of HTE estimators.
These metrics include inverse propensity weighting (IPW) validation~\citep{schuler-model18}, $ \tau $-risk~\citep{schuler-model18,nie-rlearner17}, nearest-neighbor imputation (NN-PEHE)~\citep{schwab-perfect18}, plugin validation~\citep{saito-cv19}, and counterfactual cross-validation (CFCV)~\citep{saito-cv19}.

IPW validation uses inverse propensity score weighting to a plugin estimate for the CATE, $ \tilde{\tau} $~\citep{schuler-model18}:
\begin{equation}
    \tilde{\tau}_{IPW} (\mathbf{X}_i, Y_i, T_i) = \frac{T_i}{\hat{e}(\mathbf{X}_i)}Y_i - \frac{1 - T_i}{1 - \hat{e}(\mathbf{X}_i)} Y_i.
\end{equation}
Here, $ \hat{e}(\features_i) $ is an estimate of the propensity score, which is the probability of receiving the treatment of interest (i.e., $\mathbb{E}[T_i|\mathbf{X}_i]$)
This plugin is used to approximate the mean-squared error for the estimator $ \hat{\tau} $:
\begin{equation}
    \hat{R}_{IPW}(\hat{\tau}) = \frac{1}{n} \sum_{i=1}^n \Big(\tilde{\tau}_{IPW} (\mathbf{X}_i, Y_i, T_i) - \hat{\tau}(\mathbf{X}_i) \Big)^2.
\end{equation}

The second metric $ \tau  $-risk~\citep{nie-rlearner17} takes as input an estimator $ \hat{\tau}  $ and approximates the objective defined in~\eqref{eq:cate_objective}.
To do this, two functions are defined: $ \hat{m}(\mathbf{X}_i) $ which is an outcome estimator for unit $i$ (i.e., $ \mathbb{E}[Y_i | \mathbf{X}_i] $) and $ \hat{e}(\mathbf{X}_i) $.
Both $ \hat{m} $ and $ \hat{e} $ are learned through supervised techniques and optimized for predictive accuracy~\citep{nie-rlearner17}.
The $ \tau  $-risk is defined as:
\begin{equation}
    \tau\text{-risk} = \frac{1}{N} \sum_{i=1}^N \Big( \big(Y_i-\hat{m}(\mathbf{X}_i)\big) - \big(T_i - \hat{e}(\mathbf{X}_i)\big)\hat{\tau}(\mathbf{X}_i) \Big)^2.
\end{equation}

The third metric, neighbor precision in estimation of heterogeneous effects (NN-PEHE)~\citep{schwab-perfect18} uses matching to approximate a treatment effect by substituting counterfactual outcomes from matched neighbors.
A unit $i$ is matched with one\footnote{Matching can be done with multiple units, and an aggregate can be computed.} control and one treated unit, denoted by NN$_0 (i)$ and NN$_1 (i)$.
Using the matched units, a substitute of the ground-truth treatment effect is computed as $ y(\text{NN}_1(i)) - y(\text{NN}_0(i)) $.
The NN-PEHE error, $ \hat{\epsilon}_{\text{NN-PEHE}} $ is computed as:
\begin{equation}
    \hat{\epsilon}_{\text{NN-PEHE}} = \frac{1}{N} \sum_{i=1}^n \bigg( \big(y(\text{NN}_1(i)) - y(\text{NN}_0(i)) \big) - \hat{\tau}(\mathbf{X}_i) \bigg)^2.
\end{equation}

Plugin validation~\citep{saito-cv19}, the fourth metric, uses machine learning algorithms to estimate a plugin $ \tau $, similar to IPW validation.
To get a plugin estimate, two predictions are needed, $ f_1(\mathbf{x}) $ and $ f_0(\mathbf{{x}}) $, which are predictions of the outcome when treated and not treated, respectively, using only the features.
Then, the plugin validation metric is defined as:
\begin{equation}
    \hat{R}_{plugin}(\hat{\tau}) = \frac{1}{N} \sum_{i=1}^N \bigg( (f_1(\mathbf{X}_i) - f_0(\mathbf{X}_i)) - \hat{\tau}(\mathbf{X}_i) \bigg)^2.
\end{equation}

Counterfactual cross-validation (CFCV) uses the plugin estimate of the CATE in a doubly robust formulation~\citep{saito-cv19}:
\begin{equation}
    \tilde{\tau}_{DR}(\mathbf{X}_i, Y_i, T_i) = \frac{T - \hat{e}(\mathbf{X}_i)}{\hat{e}(\mathbf{X}_i) (1 - \hat{e}(\mathbf{X}_i))} (Y_i - f_T(\mathbf{X}_i)) + f_1(\mathbf{X}_i) - f_0(\mathbf{X}_i),
\end{equation}
where $ \hat{e} $ is an estimate of the propensity score and $ f_t(\mathbf{x}) $ is an arbitrary regression function for a specific treatment value $ t $ that predicts the outcome given the set of features.
Then, the plugin estimate $ \tilde{\tau}_{DR} $ can be used in the mean-squared error estimate:
\begin{equation}
    \hat{R}_{CFCV} = \frac{1}{N} \sum_{i=1}^N \bigg( \tilde{\tau}_{DR}(\mathbf{X}_i, Y_i, T_i) - \hat{\tau}(\mathbf{X}_i) \bigg)^2
\end{equation}

Each metric is an estimate of the error of an HTE estimator and performance based on these heuristic metrics has been shown to correlate with true performance.
Since these metrics are heuristic, we explore their reliability in synthetic and semi-synthetic experiments in Section~\ref{sec:experiments}.

\section{Structural causal models for HTE estimation}\label{sec:structural_causal_models}

Causal effect identifiability in the data-driven estimation methods for counterfactual prediction and effect estimation we discussed so far %
typically rely on the ignorability assumption. However, it is not always straightforward for a practitioner to decide whether (some of) the features in the data would meet this assumption and can be used for estimation.
Including all variables in the data can be problematic in observational studies because including some variables could lead to incorrect estimation, which we discuss in more detail in Section \ref{sec:hte_estimation_with_scms}. To deal with this issue, it is helpful to know the cause-effect relationships between variables. \emph{Structural causal models} (SCMs) are graphical models that capture these relationships and can help practitioners decide which variables should and should not be used for building HTE estimators.
In this section, we provide a brief introduction to SCMs and how HTEs are represented in them.

\subsection{Structural causal models}\label{sec:scm_background}

\begin{figure}
    \centering
    \newcommand{\scmsize}{0.5}
    \begin{subfigure}[b]{\columnwidth}
        \centering
        \begin{tikzpicture}
            \node (t) at (-0.75, 0) [label=below:$T$,point];
            \node (x) at (0.0, 0.75) [label=left:\(X\),point];
            \node(y) at (0.75, 0.0) [label=below:$Y$,point];
            \node (ux) at (0.0, 1.5) [label=above:\(U_X\),point];
            \node (ut) at (-1.5, 0.75) [label=above:\(U_T\),point];
            \node (uy) at (1.5, 0.75) [label=above:\(U_Y\),point];

            \path (x) edge (t);
            \path (x) edge (y);
            \path (t) edge (y);
            \path (ux) edge (x);
            \path (uy) edge (y);
            \path (ut) edge (t);
        \end{tikzpicture}
    \end{subfigure}
    \caption{An example causal model graph.
    }\label{fig:example_scm}
\end{figure}
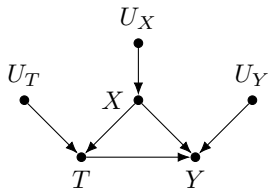

A structural causal model (SCM), $M$, consists of two sets of variables, $\mathbf{U}$ and $\mathbf{V}$, and a set of structural equations, $\mathbf{F}$, describing how values are assigned to each \emph{endogenous} variable $V^{(i)} \in \mathbf{V}$ based on the values of $\mathbf{v}$ and $\mathbf{u}$~\citep{pearl-book09}: \( v^{(i)} = f^{(i)}(\mathbf{v}, \mathbf{u}) \).
A causal graph \( G \) is a directed acyclic graph that captures the causal relationships among the variables.
Each variable, $V^{(i)}$, is represented by a node in the causal graph, an edge from $V^{(i)}$ to $V^{(j)}$ signifies that $V^{(i)}$ is a cause of $V^{(j)}$.
The edge relationship is a parent-child relationship, where the parent is a cause of the child.
The variables in $\mathbf{U}$ are considered \emph{exogenous}, and the causal graph may omit representing $\mathbf{U}$ explicitly.
Figure~\ref{fig:example_scm} shows an example of a causal graph with three exogenous variables, $ U_X, U_T, U_Y $ and three endogenous variables, $ X, T, Y $.

An example set of associated structural equations is shown in eq~\eqref{eq:see1}-\eqref{eq:see3}.
\begin{gather}
    X = U_X \label{eq:see1} \\
    T = Bern(X) + U_T \label{eq:see2}  \\
    Y = \alpha \cdot T + \beta \cdot X + U_Y \label{eq:see3}
\end{gather}
In the graph, each endogenous variable has a directed edge pointing to it from an exogenous variable (e.g., $ U_X \rightarrow X $ or $ U_X $ is a parent of $ X $).
This is represented in the structural equations as being part of the structural equation (e.g., $ U_X $ in eq~\eqref{eq:see1}).
It is assumed that every endogenous variable has a parent in $ \mathbf{U} $.
We denote the parents and children of a variable, $ X $, in the graph as $ pa(X) $ and $ ch(X) $, respectively.
Ancestors of $ X $, $ an(X) $, are variables that can reach $ X $ through a directed path, and descendants of $ X $, $ de(X) $, are variables that can be reached from $ X $ through directed paths out of $ X $.

\subsubsection{Identifying effects using SCMs}\label{sec:identifying_scms}

When the SCM underlying the dataset is known, its graph $ G $ supports the use of graphical criteria for variable selection for unbiased causal effect estimation, known as adjustment set selection~\citep{shpitser-uai06,shpitser-uai10,van-uai14,perkovic-uai15,henckel-arxiv19}.
These graphical criteria rely on the concept of $d$-separation.
\begin{definition}
    ($d$-separation) A set of $ \mathbf{Z} $ nodes is said to ``block'' a path $ p $ if either
    \begin{enumerate}
        \item[$(i)$] $ p $ contains at least one arrow-emitting node that is in $ \mathbf{Z} $ or
        \item[$(ii)$] $p$ contains at least one collision node that is outside $ \mathbf{Z} $ and has no descendant in $ \mathbf{Z} $.
    \end{enumerate}
\end{definition}
If $ \mathbf{Z} $ blocks all paths from a set of nodes $ \mathbf{X} $ to a set of nodes $ \mathbf{Y} $, it is said to ``d-separate'' $ \mathbf{X} $ and $ \mathbf{Y} $, and the conditional independence, $ \mathbf{X} \perp \mathbf{Y} \mid \mathbf{Z} $, holds~\citep{bareinboim-pnas16}.
One graphical criteria for adjustment set selection is the backdoor criterion~\citep{pearl-book09}.
The backdoor criterion aims to find a set that \emph{blocks} the paths from the treatment variable to the outcome variable in the graph, starting with an arrow into the treatment variable.
Essentially, the backdoor criterion blocks all the paths going backward out of treatment.
For example, in Figure~\ref{fig:example_scm}, there is one backdoor path, namely $ T \leftarrow X \rightarrow Y $, and so the adjustment set $ \{X\} $ sufficiently satisfies the backdoor criterion.
Several adjustment criteria have been developed, such as the adjustment criterion~\citep{shpitser-uai06,van-uai14}, generalized backdoor criterion~\citep{maathuis-annals15}, generalized adjustment criterion~\citep{perkovic-uai15}, and optimal adjustment set (O-set)~\citep{henckel-arxiv19}.
These adjustment sets defined by the various adjustment criteria can be found through \emph{identification} algorithms such as the ID algorithm~\citep{shpitser-jmlr08}, the algorithmic framework by Van der Zander et al.~\citep{van-uai14} or the pruning procedure by Henckel et al.~\citep{henckel-arxiv19}.

When an adjustment set is discovered, unbiased causal effects can be estimated by conditioning on the adjustment set.
The adjustment set can be used instead of all the features for data-driven estimation.
We discuss data-driven HTE estimation with SCMs briefly in Section~\ref{sec:hte_estimation_with_scms}.

\subsection{Heterogeneous treatment effects in SCMs}

Shpitser and Pearl provide a nonparametric estimand for HTEs for some feature assignment $C=c$, called $c$-specific effects~\citep{shpitser-uai06,pearl-smr17}.
One issue with estimating $c$-specific effects is that it requires knowledge of the SCM and which variables, C, exhibit heterogeneous effects.
When the SCM is known, the O-set~\citep{henckel-arxiv19} can be used on the learned structure since it implicitly includes all potential heterogeneity-inducing variables, while some identification algorithms may miss them. When the SCM is unknown, the identifiability of \( c \)-specific effects cannot be established.
Moreover, when the set of heterogeneity-inducing variables $\mathbf{C}$ is unknown, then all combinations of features and feature values would need to be checked for heterogeneity in effect, which is exponential in the number of valid features.
Circumventing this exponential search is one of the main reasons data-driven methods are popular in practice.

Shpitser and Pearl provide a nonparametric estimand for HTEs with respect to some feature assignment $C=c$, called $c$-specific effects~\citep{shpitser-uai06,pearl-smr17}.
There are three cases when $c$-specific effects are \emph{identifiable}, i.e., there exists an unbiased estimation for $\tau(\mathbf{c}) = E[Y(1) - Y(0) | \mathbf{C}=\mathbf{c}]$.
The first case is when $\mathbf{C}$ is observed and admissible which means that it satisfies the backdoor criterion in the causal graph~\citep{pearl-book09}.
A set of variables $\mathbf{Z}$ satisfies the backdoor criterion relative to $(T,Y)$ if no node in $\mathbf{Z}$ is a descendant of $T$ and $\mathbf{Z}$ blocks every path between $T$ and $Y$ that contains an arrow into $T$.
For example, $X$ is the only variable that is admissible in all SCMs in Figure~\ref{fig:fail_examples}, since it blocks the only backdoor path.

The second case is when $\mathbf{C}$ is part of an observed admissible set.
In this case, $\mathbf{C}$ itself is not admissible, but we can observe a set of features $\mathbf{S}$ such that $\mathbf{S} \cup \mathbf{C}$ is admissible.
In Figure~\ref{fig:marloes_scm}, $\{F\}$ could induce heterogeneity but is not admissible on its own. Adding $F$ to $\{X\}$ still satisfies the backdoor criterion making $\{X, F\}$ also admissible.
The third case for identifiable $c$-specific effects is when $\mathbf{C}$ is not part of an admissible set.
The first two cases which rely on \emph{covariate adjustment}~\citep{perkovic-jmlr18} and identifying heterogeneity-inducing variables are the basis of our work.
Estimating heterogeneous effects in the third case is beyond the scope of this paper and is reserved for future work.

One issue with estimating $c$-specific effects is that it requires knowledge of the SCM and which variables exhibit heterogeneous effects.
If the SCM is unknown, the identifiability of \( c \)-specific effects cannot be established.
If $\mathbf{C}$ is unknown, then all combinations of features and feature values would need to be checked for heterogeneity in effect, which is exponential in the number of valid features~\citep{pearl-smr17}.
Circumventing this exponential search is one of the main reasons why data-driven methods are popular in practice.

\subsubsection{Detecting latent heterogeneity}

In some cases, the observed covariates that induce heterogeneity, $C$, are not known, but there is unknown heterogeneity in effects present in the data~\citep{pearl-smr17}.
Pearl discusses how heterogeneity can be detected from statistical averages alone~\citep{pearl-smr17} when the effect of treatment on the treated ($ETT$) and the effect of treatment on the untreated ($ETU$) are identifiable.
The three cases where $ETT$ can be identified are when:
\begin{enumerate}
    \item The treatment is binary, and $E[Y_1]$ and $E[Y_0]$ are identifiable. %
    \item The treatment is arbitrary, and $E[Y_x]$ is identifiable (for all $x$) by adjustment for an admissible set of covariates.
    \item $ATE$ is identified through mediating instruments.
\end{enumerate}
The data-driven HTE estimation methods discussed in this article do not focus on the detection of latent heterogeneity, but using these cases when latent heterogeneity can be detected with data-driven estimators is an interesting future direction.

\subsubsection{Representation of HTEs in structural equations}\label{sec:htes_scms_background}

In structural equations, heterogeneous effects are represented by an \emph{interaction} or \emph{effect modification} through ancestors and parents of a variable. The DAG itself does not encode interactions but it can point to potential sources of interaction. Heterogeneity through interaction can occur with any parent of the outcome or any parent of \emph{mediating} variables~\citep{vanderweele-epid07}.
We will use linear structural equations for illustration where
the interaction is with the treatment variable.
Consider the causal model in Figure~\ref{fig:example_scm}, where the treatment variable of interest is $ T $ and the outcome of interest is $ Y $. 
There is a potential interaction between $ X $ and $ T $. 
Suppose $ Y $ is generated by the following structural equation:
\begin{equation}
    Y = \alpha \cdot T + \beta \cdot X + \gamma \cdot T \cdot X + U_Y.
\end{equation}
Here, there is a fixed effect \( \alpha \) of the treatment \( T \) and a heterogeneous effect \( \gamma \) due to the interaction between \( T \) and \( X \).
Since the effect of treatment varies with another variable, $ X $, there is a heterogeneous effect.

\begin{figure}
    \centering
    \begin{tikzpicture}
        \node (t) at (-1.5, 0) [label={below left}:$T$,point];
        \node (x) at (-1.5, 1.5) [label={above}:$X$,point];

        \node (c) at (0.0, 1.5) [label={above}:$C$,point];
        \node (d) at (0.0, 0) [label={below}:$D$,point];

        \node (f) at (1.5, 1.5) [label={above}:$F$,point];
        \node (y) at (1.5, 0.0) [label={below}:$Y$,point];

        \path (x) edge (t);

        \path (x) edge (y);
        \path(t)edge(d);
        \path(c)edge(d);
        \path (d) edge (y);

        \path (f) edge (y);
    \end{tikzpicture}
    \caption{Causal model with multiple possible sources of interactions that lead to heterogeneous treatment effects.}\label{fig:multi_interaction}
\end{figure}
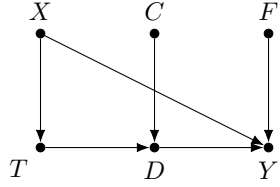

Consider another example causal model, shown in Figure~\ref{fig:multi_interaction}.
Here, there is a \emph{mediating} variable, $ D $, a variable that occurs on a path from the treatment, $ T $, to the outcome, $ Y $.
Interaction can occur with $ X, C$, or $ F $.
For example, an interaction with $ X $ and $ F $ can be shown as:
\begin{equation}
    Y = \alpha \cdot D \cdot X + \beta \cdot D \cdot F + U_Y.
\end{equation}
Since the treatment is mediated through $ D $ and $ X $ and $ F $ interact with $ D $, there will be a heterogeneous effect.
Since $ D $ is a mediator, if $ C $ interacts with $ T $ in the structural equation, then there will also be a heterogeneous effect, such as:
\begin{gather}
    D = \alpha \cdot T + \beta \cdot T \cdot C + U_D, \\
    Y = \gamma \cdot D + U_Y.
\end{gather}

\subsection{Data-driven HTE estimation using Structural Causal Models}\label{sec:hte_estimation_with_scms}

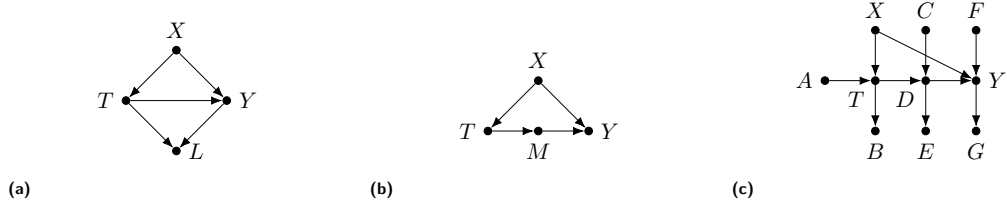
\begin{figure}
    \centering
    \newcommand{\scmsize}{0.3}
    \resizebox*{\scmsize\columnwidth}{!}{
        \begin{subfigure}[b]{0.325\columnwidth}
            \centering
            \begin{tikzpicture}
                \node (t) at (-0.75, 0) [label=left:$T$,point];
                \node (x) at (0.0, 0.75) [label=above:\(X\),point];
                \node(y) at (0.75, 0.0) [label=right:$Y$,point];
                \node(c) at (0.0, -0.75) [label=right:$L$,point];

                \path (x) edge (t);
                \path (x) edge (y);
                \path (t) edge (y);
                \path (t) edge (c);
                \path (y) edge (c);
            \end{tikzpicture}
            \caption{}\label{fig:collider_scm}
        \end{subfigure}
    }
    \resizebox*{\scmsize\columnwidth}{!}{
        \begin{subfigure}[b]{0.325\columnwidth}
            \centering
            \begin{tikzpicture}
                \node (t) at (-0.75, 0) [label=left:$T$,point];
                \node (m) at (0.0, 0) [label=below:$M$,point];
                \node (x) at (0.0, 0.75) [label=above:\(X\),point];
                \node(y) at (0.75, 0.0) [label=right:$Y$,point];

                \path (x) edge (t);
                \path (x) edge (y);
                \path (t) edge (m);
                \path (m) edge (y);
            \end{tikzpicture}
            \caption{}\label{fig:mediator_scm}
        \end{subfigure}
    }
    \resizebox*{\scmsize\columnwidth}{!}{
        \begin{subfigure}[b]{0.325\columnwidth}
            \centering
            \begin{tikzpicture}
                \node (a) at (-1.5, 0) [label=left:$A$,point];
                \node (t) at (-0.75, 0) [label={below left}:$T$,point];
                \node (b) at (-0.75, -0.75) [label={below}:$B$,point];
                \node (x) at (-0.75, 0.75) [label={above}:$X$,point];
                
                \node (c) at (0.0, 0.75) [label={above}:$C$,point];
                \node (d) at (0.0, 0) [label={below left}:$D$,point];
                \node (e) at (0.0, -0.75) [label={below}:$E$,point];
        
                \node (f) at (0.75, 0.75) [label={above}:$F$,point];
                \node (y) at (0.75, 0.0) [label={right}:$Y$,point];
                \node (g) at (0.75, -0.75) [label={below}:$G$,point];
        
                \path (a) edge (t);
                \path (x) edge (t);
                \path (t) edge (b);
                
                \path (x) edge (y);
                \path(t)edge(d);
                \path(c)edge(d);
                \path(d)edge(e);
                \path (d) edge (y);
        
                \path (f) edge (y);
                \path (y) edge (g);
            \end{tikzpicture}
            \caption{}\label{fig:marloes_scm}
        \end{subfigure}
    }
    \caption{Causal models where data-driven HTE estimation methods may perform poorly if the structure is unknown. 
    Some variables (e.g., $L$, $M$, and $E$) are not valid adjustment variables for estimating the effect of $T$ on $Y$.
    }\label{fig:fail_examples}
\end{figure}

As previously discussed, it is very important that data-driven methods use the right variables for estimation. %
Including all variables or an incorrect subset of variables would lead to incorrect effect estimation~\citep{pearl-book09}.
Incorrect variables include descendants of treatment or instrumental variables~\citep{pearl-book09,myers-aje11} (e.g., $L, M, A, B, D, E, G$ in Figure~\ref{fig:fail_examples}).
If the underlying causal model is known, then identifiability can be established, and for identifiable causal effects, an adjustment set can be discovered using the rules of do-calculus~\cite{pearl-book09}. 
Then, unbiased effects can be estimated by using the adjustment set as the features for estimation.

\subsubsection{Causal structure learning}\label{sec:structure_learning}

When the underlying causal model is unknown, it is sometimes possible to learn its structure from data~\citep{heinze-stat18,vowels-arxiv21}.
Causal structure learning algorithms can be broadly classified into constraint-based~\citep{spirtes-book00,colombo-jstor12}, score-based~\citep{chickering-jmlr02,hauser-jmlr12,nandy-annals18}, a hybrid of the two~\citep{tsamardinos-ml06}, or based on asymmetries~\citep{shimizu-jmlr06}.
The output of these methods is an equivalence class of possible structures that can explain the data. The equivalence class is represented as a partial graph or skeleton in which some of the edges may be unoriented.
Using this output, one can find an adjustment set of variables that can be used for  estimation~\citep{pearl-book09,henckel-arxiv19} and training HTE estimators.

One well-known constraint-based structure learning algorithm is the PC algorithm~\citep{spirtes-book00}.
PC conducts conditional independence tests to learn about the structure of the underlying model. It learns a completed partially directed acyclic graph (CPDAG)~\citep{chickering-jmlr02}, which is an equivalence class of directed acyclic graphs (DAG) that have uncertain edges.
PC has three main steps: (1) determining the skeleton, (2) determining v-structures, and (3), determining further edge orientations.
PC learns the skeleton by starting with a complete undirected graph.
Using a series of conditional independence tests for adjacent nodes, PC removes an edge if a conditional independence is found.
In the second step, all edges in the graph are replaced with $\circ$--$\circ$ edges.
PC considers all unshielded triples, triples $i \circ $--$\circ j$--$\circ k$ where $i$ and $k$ are not adjacent and determines whether a v-structure should exist or not.
In the final step, additional orientation rules are applied to orient as many unoriented edges as possible.
The fast causal inference (FCI) algorithm is a modification of the PC algorithm that allows hidden variables~\citep{spirtes-book00}.

A classic score-based algorithm is greedy equivalence search (GES) which also learns a CPDAG. 
Score-based algorithms typically use a likelihood score of a DAG against the data, and search for a DAG or CPDAG that yields the optimal score.
GES learns the CPDAG via a greedy forward search of edge additions, followed by a backward phase with edge deletions.
A hybrid approach based on FCI and GES has also been developed, called greedy fast causal inference (GFCI)~\citep{ogarrio-pgm16}.
Heinze-Deml et al.\ provide a survey on structure discovery methods, specifically constraint-based and score-based approaches~\citep{heinze-stat18}.

Besides constraint-based approaches and score-based approaches, continuous optimization-based methods for structure discovery have been developed due to increased compute that make learning from high-dimensional datasets feasible~\citep{ma-sr14,zheng-neurips18,kyono-neurips20,vowels-arxiv21}.
Continuous optimization-based methods recast the graph-search problem into a continuous optimization problem~\citep{vowels-arxiv21}.
Vowels et al.\ provide a survey on continuous optimization-based approaches~\citep{vowels-arxiv21}.
With the learned causal graph, an \emph{identification} algorithm can be used to find an adjustment set.
Then, this adjustment set can be used as features in an HTE estimator instead of the full feature set.

Although causal structure learning can address some problems of data-driven HTE estimation, there are some drawbacks.
First, learning the full structure can be computationally expensive for high-dimensional datasets~\citep{chickering-jmlr04,heinze-stat18,vowels-arxiv21}.
Second, causal structure learning generally cannot learn the full causal model, so learned models will have undirected edges with an ambiguous cause-effect relationship.
In addition, adjustment criteria have been developed for total causal effect estimation without heterogeneity in mind.
Therefore, adjustment sets can potentially miss heterogeneity-inducing variables, leading to incorrect HTE estimation.
In our experiments, we use randomly generated SCMs and explore how data-driven HTE estimators perform under different types of SCMs.

\section{Experiments}\label{sec:experiments}

We evaluate HTE estimators on synthetic, semi-synthetic, and real-world experiments.
We test several prominent HTE estimators in our experiments.

\subsection{Experimental setup}\label{sec:setup}

In our experiments, we compare several HTE estimators.
We employ several popular methods from recent work: Bayesian Additive Regression Trees (BART)~\citep{hill-jcgs11}, 
Causal Trees~\citep{tran-aaai19}, Causal Forest~\citep{athey-annals19},
Meta-learners (S, T, XLearners)~\citep{kunzel-pnas19}, Doubly Robust learner (DRLearner)~\citep{kennedy-arxiv20}, Balancing Neural Network~\citep{johansson-icml16}, and TARNet~\citep{shalit-icml17}.
In addition, we also use a structure learning method to find the optimal adjustment set (O-set)~\citep{henckel-arxiv19}.
We opt to use a hybrid method, greedy fast causal inference (GFCI)~\citep{ogarrio-pgm16}, and abbreviate this method as full structure learning with O-set (FSL (O-set)).
For synthetic datasets, where the causal model is known, we use an Oracle method (Known Structure) that uses the causal model for variable selection using the \emph{optimal adjustment set (O-set)}~\citep{henckel-arxiv19} and a TLearner for estimation.
We use Logistic Regression and Linear Regression for propensity scores and estimating outcomes for meta-learners on synthetic datasets.
For semi-synthetic and real-world datasets, we use Random Forest for estimating outcomes.

A related study in the literature is the work by Knaus et al.~\citep{knaus-econ21}, which investigates the performance of several heterogeneous treatment effect estimators through Monte Carlo simulations of several data generation processes. The authors compare generic modified or transformed outcome methods employing Random Forest and Lasso base learners, R-Learning, and Causal Forest. Although Knaus et al. explore some similar methods to our study, such as Causal Forest and DR-Learner, our research also considers several deep-learning, counterfactual prediction approaches, including Balancing Neural Network and TARNet. Furthermore, our study examines a distinct data generating process, specifically, how varying the underlying causal structure affects the performance of HTE estimators.

\subsection{Datasets}
We utilize three types of datasets.
The first type of dataset is a synthetic dataset.
The second type of dataset is a semi-synthetic dataset, the Infant Health and Development Program (IHDP) dataset~\citep{,brooks1992effects,hill-jcgs11}.
The final type of dataset are real-world datasets.
We use League of Legends, Cannabis, and SHARPS.

\subsubsection{Synthetic dataset}\label{sec:setup_synthetic}

We first evaluate HTE estimators on synthetic datasets generated by random structural causal models (SCMs).
As mentioned in Section~\ref{sec:hte_estimation_with_scms}, not all variables in the data should be included in HTE estimation.
A current gap in the literature is understanding how data-driven HTE estimators perform when variables should not be included due to the underlying structure, such as mediators or descendants of treatment.
Therefore, we seek to understand how estimators perform under differing structures in our synthetic experiments.
In this setting, we assume estimators do not have access to the underlying causal graph and instead use all features in the generated dataset, which mimics the real-world use of these estimators.

We generate SCMs by following a procedure similar to one for evaluating causal structure learning algorithms~\citep{heinze-stat18}.
SCMs and their corresponding datasets differ in the following characteristics: the number of variables \( d \); the probability of edge creation \( p_e \); the variance in the noise \( \sigma \); the strength of the noise term in non-source nodes \( \rho \).
In addition, we add parameters to control confounding \( \gamma \), mediation \( m \), and heterogeneity in treatment effects \( p_h, m_p \).
We initialize a set of parameters as follows:

\begin{enumerate}
    \item Number of variables: \( d \in [10, 20, 30] \)
    \item Edge probability parameter: \( p_e \in [0.1, 0.3, 0.5] \)
    \item Noise variance: \( \sigma \in [0.2, 0.4, 0.6] \)
    \item Magnitude in noise: \( \rho \in [0.1, 0.5, 0.9] \)
    \item Confounder: \( \gamma \in [\text{True, False}] \)
    \item Mediator chain length: \( m \in [0, 1, 2] \)
    \item Number of HTE inducing parents: \( p_h \in [0, 1, 2] \)
    \item HTE from mediating parents: \( m_p \in [\text{True, False}] \)
\end{enumerate}

Here, we describe how we generate synthetic datasets used in the experiments.
We first generate an adjacency matrix, \( \mathbf{A} \) over the \( d \) variables.
Assuming that the variables \( \{1, \dots, d \} \) are causally ordered.
For each pair of nodes \( i, j \), where \( i < j \) in the causal order, we sample from Bernoulli\(( p_e )\) to determine if there is an edge from \( i \) to \( j \).
With the adjacency matrix, we sample the corresponding coefficients, \( \mathbf{C} \), from Unif\(( -1, 1 )\).
Noise terms \( \epsilon_i \) are generated by sampling from a normal distribution with mean zero and variance \( \sigma \).
For a non-source node, $ i $ (nodes with parents), the noise term is multiplied by the magnitude in noise, $ \gamma $.

Next, we choose the treatment and outcome nodes based on two parameters, \( \gamma \) and \( m \). \( \gamma \) controls whether there is confounding in the generated dataset and \( m \) controls the length of the mediating chain (\( 0 \) means no mediator).
First, we select all pairs of nodes with a \emph{directed} \( m \) hop path.
Then, if \( \gamma \) is True, we filter all pairs that do not contain a \emph{backdoor path}.
Otherwise, we filter out pairs that contain a backdoor path.
Finally, we sample a pair of nodes from the filtered set to obtain the treatment and outcome nodes.
If no pairs satisfy the parameter setting (e.g., no confounding), the graph is resampled up to 100 times.
Otherwise, we discard the parameter setting.

A source node \( i \) (node with no parents) takes the value of the sampled noise terms, \( \epsilon_i \).
For a non-source node, \( i \), we consider several additive factors.
First is the base term from the coefficient matrix: \( C_{pa(i), i} X_{pa(i)} \), where \( C_{pa(i), i} \) are the coefficients from \( pa(i) \) to \( i \) and \( X_{pa(i)} \) are the values generated for \( pa(i) \).
Next are the interaction terms, which are determined by the mediating chain length \(m\), number of HTE-inducing parents \( p_h \), and whether there is HTE from parents of mediators \( m_p \).
For mediators, we have: \( \mathbbm{1}_{m}(i) \mathbbm{1}_{m_p} X_{pa_m(i)} X_{pa_n(i)} \), where \( \mathbbm{1}_{m} \) is an indicator for if \( i \) is a mediator, \( \mathbbm{1}_{m_p} \) is an indicator for if there is HTE induced from mediator parents, and \( pa_m \) and \( pa_n \) represent mediating and non-mediating parents.
For outcomes, the interaction is defined similarly: \( \mathbbm{1}_{y}(i) X_{pa_m(i)} X_{pa_n(i)} \), where \( \mathbbm{1}_{y}(i) \) indicates whether the node is the outcome node.
The number of HTE parents determines how many non-mediator parents (\( pa_n \)) interact with the treatment.
A value of zero means there is no heterogeneity.
So the final data-generating equation is:
\begin{align}
    X_i & = C_{pa(i), i} X_{pa(i)} 
    \nonumber \\
    & + \mathbbm{1}_{m}(i) \mathbbm{1}_{m_p} X_{pa_m(i)} X_{pa_n(i)} 
    \nonumber \\
    & + \mathbbm{1}_{y}(i) X_{pa_m(i)} X_{pa_n(i)} + \epsilon_i
\end{align}

For the $X_i$ that is selected to be a treatment variable, we apply a sigmoid function to constrain the variable in the $[0,1]$ range. 
Then, for each example, $i$, we sample a value, $t_i$ uniformly from $[0,1]$.
If $sigmoid(X_i) \geq t_i$, we set $X_i = T_i = 1$.
Otherwise, $T_i = 0$.

For each parameter, we randomly sample from all other parameter values \( 500 \) times to generate SCMs.
For each of the $11,000$ SCMs, we create one dataset of size \( 10,000 \) and one dataset of size $1,000$.

\subsubsection{IHDP dataset}
The covariates in the Infant Health and Development Program (IHDP) dataset come from a randomized experiment studying the effects of specialist home visits on future cognitive test scores~\citep{brooks1992effects, hill-jcgs11}.
Outcomes can be simulated using the covariates.
The dataset size is 747. There are 6 continuous features and 19 binary features, with a continuous outcome variable.
We simulate 1,000 instances of simulated outcomes and average over all results.
We use the simulated outcome implemented as setting “A” in the NPCI package~\citep{doriegithub}, the same as in~\citep{shalit-icml17}. Outcomes are generated using the JustCause package~\citep{inovex}.

\subsubsection{Real-world datasets}

We additionally explore three real-world datasets.
The first dataset is a League of Legends (LoL) dataset~\citep{he-fdg21}.
LoL is one of the most popular multiplayer online battle arena (MOBA) games.
Players compete in teams to capture the opposing team's base.
Riot Games, the developer of LoL, regularly updates the game's software through \emph{patches}.
The LoL dataset consists of 1.2 million unique players in 437,000 matches.
The dataset contains information about players and matches for different versions of the game.
We use 344 features for the HTE estimators that consist of a mix of continuous and binary values.
Here, we use patches as the binary treatment variable and measure how it affects the number of \emph{kills} on a team level.

\emph{Cannabis} is a dataset consisting of tweets about the e-cigarette Juul and cannabis-related topics~\citep{adhikari-icwsm21}.
English tweets related to the e-cigarette Juul were collected from 2016 to 2018.
For the users who mention Juul, cannabis-related tweets were collected from 2014 to 2018.
In this dataset, stances towards Juul and cannabis in tweets are crowdsourced, and a model is trained for stance detection.
In this work, we are interested in how users' stance on Juul affects their stance on cannabis.
To do this, we find users who have Juul-related tweets occurring before a cannabis-related tweet and detect their stance on Juul and cannabis using the classifier from~\citep{adhikari-icwsm21}, where features are the word embeddings for tweets using BERT~\citep{devlin-nacl19}.
The treatment variable of interest is a user's stance on Juul, in favor (1) or against (0).
The outcome variable is their stance on cannabis, in favor (1) or against (0).
We consider the Empath~\citep{fast-chi16} categories of the tweets as features, which have 194 continuous variables.
Since this dataset contains only one group, we bootstrap the estimators to compute the standard deviation.

The third dataset is about a study of interventions that may improve performance among undergraduates at the University of Illinois at Urbana Champaign (UIUC): strengthening human adaptive reasoning and problem-solving (SHARPS).
In this dataset, students were given a battery of cognitive, physical, and diet evaluations.
In subsequent weeks, they were given one of four interventions: no training (active control), fitness training, fitness and cognitive training, and fitness, cognitive, and mindfulness training.
Since most HTE estimators can only handle binary treatment variables, we measure the effect of fitness training compared to active training.
The outcome variable we are interested in is the change in the number of correct answers on a digit-symbol substitution task.
There are 239 samples with 316 continuous features. 
To compute standard deviation, we bootstrap estimators.

\subsection{Evaluation}

When the ground truth is known, a common metric is the \emph{precision in estimation of heterogeneous effects} (PEHE)~\cite{hill-jcgs11}:
\begin{equation}
    PEHE = \sqrt{\frac{1}{n} \sum_i^n \Big( (Y_i(1) - Y_i(0)) - E[Y_i(1) - Y_i(0) \mid \mathbf{X}_i = \mathbf{x}_i] \Big)^2},
\end{equation}
where the expectation is the prediction of an HTE estimator. 
This is the root mean squared error (RMSE) of the true effect and the predicted effect.

In practice, we may be more interested in how well an HTE estimator can predict whether an individual will benefit from treatment or not.
For synthetic datasets, we compute an \emph{accuracy of decisions}.
For an individual, $ i $, let the true individual effect be $ \tau_i $, and the predicted effect be $ \hat{\tau}_i $.
Let the indicator $ \mathbbm{1}\big[ sign(\tau_i) = sign(\hat{\tau}_i) \big] $ determine if the sign of the predicted effect matches the sign of the true effect.
The accuracy is defined as:
\begin{equation}
    Acc = \frac{1}{N} \sum_{i}^{N} \mathbbm{1}[sign(\tau_i) = sign(\hat{\tau}_i)].
\end{equation}

If the ground truth is unknown, several metrics for heuristically evaluating HTE estimators have been proposed~\citep{saito-cv19,schuler-model18}.
Therefore, we additionally report the $ \tau  $-risk for evaluation:
\begin{equation}
    \tau\text{-risk} = \frac{1}{n} = \sum_i^n \Big( (Y_i - \hat{m}(\mathbf{X}_i)) - (T_i - \hat{p}(\mathbf{X}_i)) \hat{\tau}(\mathbf{X_i}) \Big)^2,
\end{equation}

When comparing estimators across different parameter settings and SCMs, we also rank all methods based on their error for a particular SCM\@.
Then we average the rank of each method across SCMs and report on the average rank.
To study performance under specific settings, we also report on average ranks for subsets of parameter settings.

\subsection{Synthetic dataset evaluation}\label{sec:synthetic_eval}

We first investigate how HTE estimation methods perform under different parameter settings (and thus SCMs). 
Then we investigate the effect that causal feature selection has on HTE estimation.

\subsubsection{Multidimensional scaling}\label{sec:synthetic_mds}

\begin{figure*}[t]
    \centering
    \includegraphics[width=\linewidth]{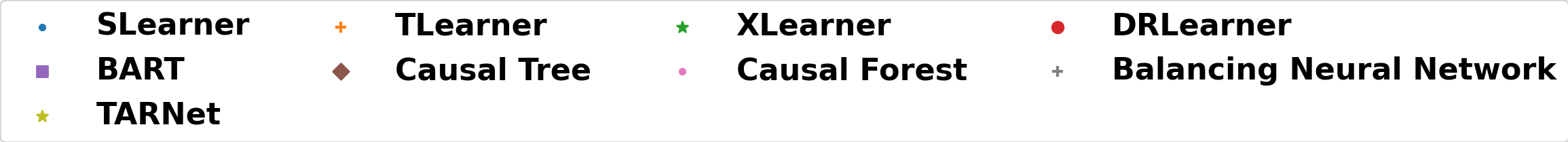}
    \includegraphics[width=0.7\textwidth]{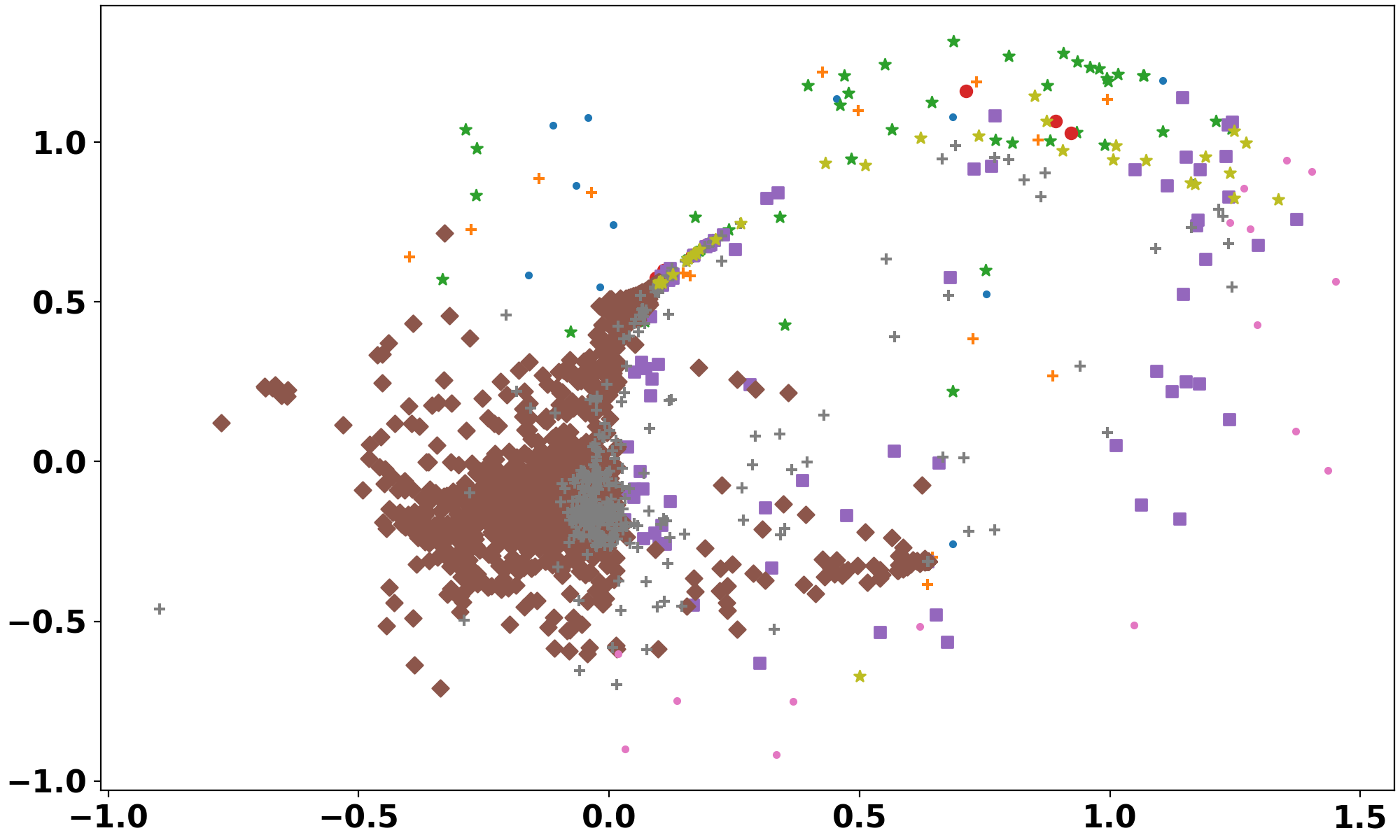}
    \caption{A multidimensional scaling (MDS) visualization of the simulated datasets.
    The distance between two simulation settings is the correlation distance between the RMSE of both simulated datasets. Each dataset is shown as a point with color and shape coding of the best-performing method.}\label{fig:mds_correlation}
\end{figure*}

For each method and SCM, we compute the RMSE for the predicted effect.
This yields a (number of methods) $ \times $ (number of SCMs) matrix with RMSE values.
The correlation between pairs of SCM results is the similarity between two datasets.
Figure~\ref{fig:mds_correlation} shows a Multidimensional scaling (MDS) plot based on the correlation between SCM datasets.
The goal of the MDS plot is to visualize the similarity in performance of methods based on different SCM settings.
If two points are close together, then all learners in those experiments performed similarly.
The color and shape of a point represent the method that performed the best (lowest RMSE) for that SCM\@.

From this Figure, Causal Tree has the best performance in most settings, followed by Balancing Neural Network and BART.
Interestingly, Causal Tree and Balancing Neural Network have much overlap, meaning that the settings in which they perform the best have similar RMSE across all learners.
BART and other meta-learners occupy more space on the opposite side of Causal Tree.

\begin{table*}[t]
    \centering
    \resizebox*{\linewidth}{!}{
        \begin{tabular}{ P{2.2cm} *{9}{| M{1.3cm}}  N}
            & SLearner & TLearner & XLearner & DRLearner & BART & Causal Tree & Causal Forest & Balancing Neural Network & TARNet \\
            \midrule
            SLearner & 0.00 & 0.00 & 0.34 & 46.69 & 68.45 & 68.80 & 18.30 & 30.28 & 37.61 \\
            \midrule
            TLearner & 49.88 & 0.00 & 0.14 & 55.09 & 63.28 & 69.86 & 24.29 & 34.43 & 44.13 \\
            \midrule
            XLearner & 50.12 & 0.34 & 0.00 & 55.09 & 63.17 & 69.91 & 24.14 & 34.43 & 43.89 \\
            \midrule
            DRLearner & 28.49 & 13.57 & 13.52 & 0.00 & 70.88 & 64.55 & 12.17 & 20.28 & 27.04 \\
            \midrule
            BART & 63.72 & 62.84 & 62.95 & 70.88 & 0.00 & 49.72 & 27.65 & 42.79 & 64.10 \\
            \midrule
            Causal Tree & 79.23 & 78.55 & 78.55 & 83.47 & 48.35 & 0.00 & 77.15 & 75.27 & 80.72 \\
            \midrule
            Causal Forest & 58.91 & 44.91 & 44.86 & 62.10 & 42.47 & 71.06 & 0.00 & 41.28 & 47.90 \\
            \midrule
            Balancing Neural Network & 58.91 & 51.71 & 51.86 & 69.48 & 45.28 & 72.80 & 43.31 & 0.00 & 58.14 \\
            \midrule
            TARNet & 41.24 & 28.83 & 29.02 & 49.93 & 61.67 & 67.88 & 20.23 & 20.81 & 0.00 \\
        \end{tabular}
    }
    \caption{Pairwise comparison between HTE estimation methods. Each row shows the percentage of datasets where the row method performs better than the column method by more than 5\%. For example, SLearner never beats TLearner by more than 5\% and TLearner beats SLearner by more than 5\% in 49.88\% of scenarios.}\label{tab:pairwise}
 \end{table*}

\subsubsection{Pairwise comparison}

Next, we investigate if there are methods that beat others consistently.
We compute the percentage of wins in Table~\ref{tab:pairwise} for which there is a 5\% decrease in RMSE\@.
For example, SLearner beats XLearner by a 5\% margin in only 0.34\% of scenarios, while XLearner beats SLearner by a 5\% margin in 50.12\% of scenarios.
From these results, there does not appear to be an overly dominant method.
For example, Causal Forest generally beats DRLearner by the margin but does not generally beat by Causal Tree or Balancing Neural Network.

From this Table, there is no clear dominant method, but there are methods which are dominated by others.
For example, SLearner generally does not perform well, especially its other meta-learning counterparts.
Causal Tree has an overall high percentage of wins but is also beaten by other methods at a relatively high percentage.
This suggests methods need to be carefully chosen for different settings and assumptions.

\subsubsection{Performance in different SCMs}

\begin{table*}[t]
    \centering
    \resizebox*{\linewidth}{!}{
	\renewcommand{\minval}{1.0}
	\renewcommand{\maxval}{10.0}
	\begin{tabular}{ N P{3.0cm} | *{11}{M{1.9cm}} N }
		& &  Mediator Chain Length 0 &  Mediator Chain Length 1 &  Mediator Chain Length 2 &  Contains Confounding &  No Confounding &  0 HTE Parents &  1 HTE Parent &  2 HTE Parents &  10 Variables &  20 Variables &  30 Variables  \\
		\toprule
		& SLearner & \gradient{8.63}\( \pm \) 1.74 & \gradient{9.43}\( \pm \) 1.86 & \gradient{9.50}\( \pm \) 1.95 & \gradient{8.94}\( \pm \) 1.97 & \gradient{9.29}\( \pm \) 1.80 & \gradient{9.49}\( \pm \) 1.72 & \gradient{8.99}\( \pm \) 1.93 & \gradient{8.96}\( \pm \) 1.92 & \gradient{8.96}\( \pm \) 1.92 & \gradient{9.11}\( \pm \) 1.94 & \gradient{9.31}\( \pm \) 1.79& \\[10pt] 
		& TLearner & \gradient{6.36}\( \pm \) 1.66 & \gradient{6.93}\( \pm \) 1.79 & \gradient{6.93}\( \pm \) 1.85 & \gradient{6.51}\( \pm \) 1.97 & \gradient{6.86}\( \pm \) 1.61 & \gradient{7.04}\( \pm \) 1.62 & \gradient{6.56}\( \pm \) 1.83 & \gradient{6.55}\( \pm \) 1.83 & \gradient{6.51}\( \pm \) 1.79 & \gradient{6.77}\( \pm \) 1.82 & \gradient{6.81}\( \pm \) 1.72& \\[10pt] 
		& XLearner & \gradient{6.74}\( \pm \) 2.03 & \gradient{7.72}\( \pm \) 2.00 & \gradient{7.20}\( \pm \) 2.13 & \gradient{6.88}\( \pm \) 2.28 & \gradient{7.45}\( \pm \) 1.91 & \gradient{7.61}\( \pm \) 2.04 & \gradient{7.03}\( \pm \) 2.06 & \gradient{7.01}\( \pm \) 2.11 & \gradient{7.04}\( \pm \) 2.22 & \gradient{7.41}\( \pm \) 2.06 & \gradient{7.18}\( \pm \) 2.01& \\[10pt] 
		& DRLearner & \gradient{7.65}\( \pm \) 1.96 & \gradient{8.16}\( \pm \) 1.70 & \gradient{7.78}\( \pm \) 2.03 & \gradient{7.38}\( \pm \) 2.24 & \gradient{8.20}\( \pm \) 1.54 & \gradient{7.88}\( \pm \) 1.73 & \gradient{7.90}\( \pm \) 1.97 & \gradient{7.81}\( \pm \) 2.01 & \gradient{7.56}\( \pm \) 1.97 & \gradient{7.98}\( \pm \) 1.77 & \gradient{7.98}\( \pm \) 1.94& \\[10pt] 
		& BART & \gradient{5.80}\( \pm \) 2.43 & \gradient{5.32}\( \pm \) 2.07 & \gradient{5.78}\( \pm \) 2.37 & \gradient{5.11}\( \pm \) 2.50 & \gradient{5.97}\( \pm \) 2.08 & \gradient{5.59}\( \pm \) 2.21 & \gradient{5.65}\( \pm \) 2.33 & \gradient{5.63}\( \pm \) 2.36 & \gradient{5.84}\( \pm \) 2.41 & \gradient{5.68}\( \pm \) 2.37 & \gradient{5.43}\( \pm \) 2.16& \\[10pt] 
		& Causal Tree & \gradient{3.96}\( \pm \) 3.43 & \gradient{3.39}\( \pm \) 2.68 & \gradient{3.78}\( \pm \) 3.16 & \gradient{5.40}\( \pm \) 3.89 &\bftab \gradient{2.53}\( \pm \) 1.59 & \gradient{3.47}\( \pm \) 3.21 & \gradient{3.77}\( \pm \) 3.02 & \gradient{3.89}\( \pm \) 3.10 & \gradient{3.69}\( \pm \) 3.09 & \gradient{3.48}\( \pm \) 2.98 & \gradient{3.89}\( \pm \) 3.22& \\[10pt] 
		& Causal Forest & \gradient{9.67}\( \pm \) 2.36 & \gradient{8.24}\( \pm \) 2.52 & \gradient{7.94}\( \pm \) 2.54 & \gradient{8.15}\( \pm \) 2.92 & \gradient{9.09}\( \pm \) 2.24 & \gradient{8.52}\( \pm \) 2.62 & \gradient{8.79}\( \pm \) 2.55 & \gradient{8.80}\( \pm \) 2.56 & \gradient{9.14}\( \pm \) 2.32 & \gradient{8.55}\( \pm \) 2.63 & \gradient{8.52}\( \pm \) 2.67& \\[10pt] 
		& Balancing Neural Network & \gradient{4.64}\( \pm \) 2.26 & \gradient{4.48}\( \pm \) 1.98 & \gradient{4.53}\( \pm \) 2.14 & \gradient{4.80}\( \pm \) 2.61 & \gradient{4.38}\( \pm \) 1.70 & \gradient{4.37}\( \pm \) 2.04 & \gradient{4.61}\( \pm \) 2.14 & \gradient{4.68}\( \pm \) 2.20 & \gradient{4.78}\( \pm \) 2.08 & \gradient{4.47}\( \pm \) 2.16 & \gradient{4.46}\( \pm \) 2.14& \\[10pt] 
		& TARNet & \gradient{7.57}\( \pm \) 2.41 & \gradient{7.62}\( \pm \) 2.12 & \gradient{7.13}\( \pm \) 2.30 & \gradient{6.76}\( \pm \) 2.50 & \gradient{7.97}\( \pm \) 1.99 & \gradient{7.17}\( \pm \) 2.10 & \gradient{7.64}\( \pm \) 2.35 & \gradient{7.58}\( \pm \) 2.37 & \gradient{7.58}\( \pm \) 2.38 & \gradient{7.55}\( \pm \) 2.19 & \gradient{7.33}\( \pm \) 2.29& \\[10pt] 
		& FSL (O-set) &\bftab \gradient{3.03}\( \pm \) 1.52 &\bftab \gradient{2.98}\( \pm \) 1.66 &\bftab \gradient{3.16}\( \pm \) 2.16 &\bftab \gradient{3.48}\( \pm \) 2.24 & \gradient{2.75}\( \pm \) 1.24 &\bftab \gradient{2.97}\( \pm \) 1.57 &\bftab \gradient{3.08}\( \pm \) 1.83 &\bftab \gradient{3.08}\( \pm \) 1.85 &\bftab \gradient{2.75}\( \pm \) 1.79 &\bftab \gradient{3.20}\( \pm \) 1.63 &\bftab \gradient{3.14}\( \pm \) 1.81& \\[10pt] 
		\midrule
		& Known Structure & \gradient{1.95} \( \pm \) 1.53 & \gradient{1.72} \( \pm \) 1.63 & \gradient{2.28} \( \pm \) 2.39 & \gradient{2.59} \( \pm \) 2.58 & \gradient{1.52} \( \pm \) 0.79 & \gradient{1.88} \( \pm \) 1.27 & \gradient{1.99} \( \pm \) 2.05 & \gradient{2.00} \( \pm \) 2.07 & \gradient{2.15} \( \pm \) 1.80 & \gradient{1.80} \( \pm \) 1.62 & \gradient{1.95} \( \pm \) 2.00 \\[10pt] 
	\end{tabular}	
    }
    \caption{Average rank for different HTE estimation methods under varying experimental settings with sample size of 10000. 
    Each column shows a fixed parameter value, and averaged ranks over all other settings.
	}\label{tab:mse_rankings}
\end{table*}

In order to investigate how HTE estimators perform in different types of SCMs with $10,000$ samples, we show the average rank under different parameter settings in Table~\ref{tab:mse_rankings} and the average RMSE in Table~\ref{tab:rmse_table}.
The Oracle method (Known Structure) is at the bottom of each table.
Bolded results indicate the best rank in the column (ignoring the Oracle).
In terms of ranking, structure learning (FSL) achieves the lowest average rank over most datasets.
Additionally, since some ranks are close, there is no clear winner among closely ranked methods, which supports the pairwise comparison between HTE estimation methods in Table~\ref{tab:pairwise}.

To investigate how different methods perform when the training set varies in size, we conduct an experiment with 1,000 samples and present the results of average ranking in Table~\ref{tab:mse_rankings_1000}. Unlike previous work~\cite{knaus-econ21} which showed that the relative performance of the method depended on the sample size, we didn't find such differences here, likely due to the different experimental setup. %

\begin{table*}[t]
    \centering
    \resizebox*{\linewidth}{!}{
		\renewcommand{\minval}{1.0}
		\renewcommand{\maxval}{10.0}
		\begin{tabular}{ N P{3.0cm} | *{11}{M{1.9cm}} N }
			& &  Mediator Chain Length 0 &  Mediator Chain Length 1 &  Mediator Chain Length 2 &  Contains Confounding &  No Confounding &  0 HTE Parents &  1 HTE Parent &  2 HTE Parents &  10 Variables &  20 Variables &  30 Variables  \\
			\toprule
			& SLearner & \gradient{8.19}\( \pm \) 1.85 & \gradient{9.04}\( \pm \) 1.95 & \gradient{9.20}\( \pm \) 1.97 & \gradient{8.52}\( \pm \) 2.06 & \gradient{8.93}\( \pm \) 1.88 & \gradient{8.92}\( \pm \) 1.92 & \gradient{8.72}\( \pm \) 1.98 & \gradient{8.65}\( \pm \) 1.99 & \gradient{8.54}\( \pm \) 2.00 & \gradient{8.78}\( \pm \) 1.94 & \gradient{8.89}\( \pm \) 1.95& \\[10pt] 
			& TLearner & \gradient{6.36}\( \pm \) 1.68 & \gradient{6.99}\( \pm \) 1.81 & \gradient{7.08}\( \pm \) 1.87 & \gradient{6.50}\( \pm \) 2.00 & \gradient{6.97}\( \pm \) 1.63 & \gradient{6.96}\( \pm \) 1.70 & \gradient{6.71}\( \pm \) 1.83 & \gradient{6.66}\( \pm \) 1.87 & \gradient{6.54}\( \pm \) 1.83 & \gradient{6.85}\( \pm \) 1.78 & \gradient{6.88}\( \pm \) 1.80& \\[10pt] 
			& XLearner & \gradient{6.94}\( \pm \) 2.16 & \gradient{7.97}\( \pm \) 2.12 & \gradient{7.71}\( \pm \) 2.21 & \gradient{7.00}\( \pm \) 2.44 & \gradient{7.86}\( \pm \) 1.96 & \gradient{7.78}\( \pm \) 2.13 & \gradient{7.39}\( \pm \) 2.21 & \gradient{7.36}\( \pm \) 2.25 & \gradient{7.30}\( \pm \) 2.37 & \gradient{7.64}\( \pm \) 2.10 & \gradient{7.55}\( \pm \) 2.17& \\[10pt] 
			& DRLearner & \gradient{8.23}\( \pm \) 1.98 & \gradient{8.32}\( \pm \) 1.99 & \gradient{8.14}\( \pm \) 2.24 & \gradient{7.68}\( \pm \) 2.42 & \gradient{8.63}\( \pm \) 1.64 & \gradient{8.37}\( \pm \) 1.95 & \gradient{8.16}\( \pm \) 2.07 & \gradient{8.19}\( \pm \) 2.13 & \gradient{8.05}\( \pm \) 2.06 & \gradient{8.26}\( \pm \) 1.95 & \gradient{8.36}\( \pm \) 2.11& \\[10pt] 
			& BART & \gradient{5.53}\( \pm \) 2.28 & \gradient{5.30}\( \pm \) 2.04 & \gradient{5.61}\( \pm \) 2.26 & \gradient{5.00}\( \pm \) 2.41 & \gradient{5.79}\( \pm \) 1.97 & \gradient{5.48}\( \pm \) 2.09 & \gradient{5.48}\( \pm \) 2.22 & \gradient{5.46}\( \pm \) 2.27 & \gradient{5.52}\( \pm \) 2.21 & \gradient{5.54}\( \pm \) 2.25 & \gradient{5.38}\( \pm \) 2.13& \\[10pt] 
			& Causal Tree & \gradient{3.74}\( \pm \) 3.23 & \gradient{3.36}\( \pm \) 2.65 & \gradient{3.70}\( \pm \) 2.99 & \gradient{5.04}\( \pm \) 3.75 &\bftab \gradient{2.59}\( \pm \) 1.65 & \gradient{3.32}\( \pm \) 3.03 & \gradient{3.69}\( \pm \) 2.89 & \gradient{3.78}\( \pm \) 2.97 & \gradient{3.37}\( \pm \) 2.76 & \gradient{3.62}\( \pm \) 2.98 & \gradient{3.73}\( \pm \) 3.09& \\[10pt] 
			& Causal Forest & \gradient{9.31}\( \pm \) 2.53 & \gradient{7.47}\( \pm \) 2.50 & \gradient{6.78}\( \pm \) 2.48 & \gradient{7.77}\( \pm \) 3.05 & \gradient{8.15}\( \pm \) 2.46 & \gradient{7.74}\( \pm \) 2.77 & \gradient{8.10}\( \pm \) 2.71 & \gradient{8.13}\( \pm \) 2.69 & \gradient{8.37}\( \pm \) 2.66 & \gradient{7.93}\( \pm \) 2.76 & \gradient{7.77}\( \pm \) 2.71& \\[10pt] 
			& Balancing Neural Network & \gradient{4.74}\( \pm \) 2.39 & \gradient{4.39}\( \pm \) 2.08 & \gradient{4.47}\( \pm \) 2.27 & \gradient{4.88}\( \pm \) 2.73 & \gradient{4.31}\( \pm \) 1.82 & \gradient{4.47}\( \pm \) 2.24 & \gradient{4.58}\( \pm \) 2.24 & \gradient{4.57}\( \pm \) 2.28 & \gradient{5.01}\( \pm \) 2.35 & \gradient{4.33}\( \pm \) 2.22 & \gradient{4.38}\( \pm \) 2.17& \\[10pt] 
			& TARNet & \gradient{7.86}\( \pm \) 2.40 & \gradient{8.08}\( \pm \) 2.22 & \gradient{7.62}\( \pm \) 2.42 & \gradient{7.13}\( \pm \) 2.58 & \gradient{8.40}\( \pm \) 2.03 & \gradient{7.83}\( \pm \) 2.23 & \gradient{7.93}\( \pm \) 2.41 & \gradient{7.86}\( \pm \) 2.41 & \gradient{8.18}\( \pm \) 2.33 & \gradient{7.90}\( \pm \) 2.31 & \gradient{7.65}\( \pm \) 2.38& \\[10pt] 
			& FSL (O-set) &\bftab \gradient{3.02}\( \pm \) 1.59 &\bftab \gradient{3.14}\( \pm \) 1.86 &\bftab \gradient{3.24}\( \pm \) 2.21 &\bftab \gradient{3.60}\( \pm \) 2.39 & \gradient{2.78}\( \pm \) 1.30 &\bftab \gradient{3.04}\( \pm \) 1.76 &\bftab \gradient{3.14}\( \pm \) 1.90 &\bftab \gradient{3.17}\( \pm \) 1.94 &\bftab \gradient{2.83}\( \pm \) 1.87 &\bftab \gradient{3.21}\( \pm \) 1.74 &\bftab \gradient{3.24}\( \pm \) 1.94& \\[10pt] 
			\midrule
			& Known Structure & \gradient{2.08} \( \pm \) 1.68 & \gradient{1.95} \( \pm \) 1.91 & \gradient{2.46} \( \pm \) 2.55 & \gradient{2.90} \( \pm \) 2.76 & \gradient{1.60} \( \pm \) 1.00 & \gradient{2.11} \( \pm \) 1.59 & \gradient{2.11} \( \pm \) 2.21 & \gradient{2.17} \( \pm \) 2.22 & \gradient{2.29} \( \pm \) 1.97 & \gradient{1.93} \( \pm \) 1.83 & \gradient{2.17} \( \pm \) 2.20 \\[10pt] 
		\end{tabular}
	}
    \caption{Average rank for different HTE estimation methods under varying experimental settings with sample size of 1000. 
    Each column shows a fixed parameter value, and averaged ranks over all other settings.
	}\label{tab:mse_rankings_1000}
\end{table*}

\begin{table*}[t]
    \centering
    \resizebox*{\linewidth}{!}{
		\renewcommand{\minval}{0.0}
		\renewcommand{\maxval}{3.0}
		\begin{tabular}{ N P{3.0cm} | *{11}{M{1.9cm}} N }
			& &  Mediator Chain Length 0 &  Mediator Chain Length 1 &  Mediator Chain Length 2 &  Contains Confounding &  No Confounding &  0 HTE Parents &  1 HTE Parent &  2 HTE Parents &  10 Variables &  20 Variables &  30 Variables  \\
			\toprule
			& SLearner & \gradient{1.52}\( \pm \) 1.40 & \gradient{2.35}\( \pm \) 1.32 & \gradient{2.39}\( \pm \) 1.39 & \gradient{1.55}\( \pm \) 1.18 & \gradient{2.40}\( \pm \) 1.49 & \gradient{1.96}\( \pm \) 1.43 & \gradient{2.09}\( \pm \) 1.42 & \gradient{2.08}\( \pm \) 1.44 & \gradient{2.20}\( \pm \) 1.67 & \gradient{1.99}\( \pm \) 1.31 & \gradient{1.99}\( \pm \) 1.33& \\[10pt] 
			& TLearner & \gradient{1.38}\( \pm \) 1.27 & \gradient{2.14}\( \pm \) 1.20 & \gradient{2.17}\( \pm \) 1.27 & \gradient{1.41}\( \pm \) 1.07 & \gradient{2.18}\( \pm \) 1.35 & \gradient{1.78}\( \pm \) 1.30 & \gradient{1.90}\( \pm \) 1.28 & \gradient{1.90}\( \pm \) 1.31 & \gradient{2.00}\( \pm \) 1.52 & \gradient{1.81}\( \pm \) 1.19 & \gradient{1.80}\( \pm \) 1.21& \\[10pt] 
			& XLearner & \gradient{1.41}\( \pm \) 1.30 & \gradient{2.17}\( \pm \) 1.22 & \gradient{2.18}\( \pm \) 1.28 & \gradient{1.42}\( \pm \) 1.08 & \gradient{2.21}\( \pm \) 1.37 & \gradient{1.80}\( \pm \) 1.32 & \gradient{1.93}\( \pm \) 1.30 & \gradient{1.92}\( \pm \) 1.33 & \gradient{2.03}\( \pm \) 1.55 & \gradient{1.84}\( \pm \) 1.21 & \gradient{1.82}\( \pm \) 1.22& \\[10pt] 
			& DRLearner & \gradient{58.19}\( \pm \) 1670.68 & \gradient{10.55}\( \pm \) 170.89 & \gradient{2.26}\( \pm \) 1.45 & \gradient{6.88}\( \pm \) 148.17 & \gradient{40.17}\( \pm \) 1343.03 & \gradient{1.89}\( \pm \) 1.57 & \gradient{66.68}\( \pm \) 1775.95 & \gradient{10.42}\( \pm \) 196.76 & \gradient{2.10}\( \pm \) 1.51 & \gradient{2.22}\( \pm \) 10.89 & \gradient{61.00}\( \pm \) 1609.51& \\[10pt] 
			& BART & \gradient{1.11}\( \pm \) 0.91 & \gradient{1.61}\( \pm \) 0.93 & \gradient{1.76}\( \pm \) 1.00 & \gradient{1.16}\( \pm \) 0.86 & \gradient{1.66}\( \pm \) 1.01 & \gradient{1.33}\( \pm \) 0.93 & \gradient{1.53}\( \pm \) 1.00 & \gradient{1.51}\( \pm \) 1.01 & \gradient{1.39}\( \pm \) 1.02 & \gradient{1.46}\( \pm \) 0.91 & \gradient{1.50}\( \pm \) 1.01& \\[10pt] 
			& Causal Tree &\bftab \gradient{0.44}\( \pm \) 0.56 & \gradient{0.69}\( \pm \) 0.67 & \gradient{0.75}\( \pm \) 0.76 & \gradient{0.85}\( \pm \) 0.59 & \gradient{0.91}\( \pm \) 0.67 &\bftab \gradient{0.30}\( \pm \) 0.47 & \gradient{0.75}\( \pm \) 0.69 & \gradient{1.03}\( \pm \) 0.71 & \gradient{0.57}\( \pm \) 0.66 &\bftab \gradient{0.61}\( \pm \) 0.71 & \gradient{0.94}\( \pm \) 0.63& \\[10pt] 
			& Causal Forest & \gradient{2.16}\( \pm \) 1.13 & \gradient{2.22}\( \pm \) 1.25 & \gradient{2.23}\( \pm \) 1.30 & \gradient{1.66}\( \pm \) 1.07 & \gradient{2.58}\( \pm \) 1.18 & \gradient{2.10}\( \pm \) 1.24 & \gradient{2.25}\( \pm \) 1.20 & \gradient{2.25}\( \pm \) 1.21 & \gradient{2.64}\( \pm \) 1.18 & \gradient{2.11}\( \pm \) 1.16 & \gradient{1.97}\( \pm \) 1.21& \\[10pt] 
			& Balancing Neural Network & \gradient{0.66}\( \pm \) 0.60 & \gradient{1.07}\( \pm \) 0.73 & \gradient{1.10}\( \pm \) 0.76 & \gradient{0.94}\( \pm \) 0.62 &\bftab \gradient{0.44}\( \pm \) 0.79 & \gradient{0.69}\( \pm \) 0.61 & \gradient{1.04}\( \pm \) 0.74 & \gradient{0.78}\( \pm \) 0.75 & \gradient{0.92}\( \pm \) 0.79 & \gradient{0.90}\( \pm \) 0.72 &\bftab \gradient{0.64}\( \pm \) 0.67& \\[10pt] 
			& TARNet & \gradient{1.58}\( \pm \) 1.23 & \gradient{2.16}\( \pm \) 1.19 & \gradient{2.17}\( \pm \) 1.26 & \gradient{1.40}\( \pm \) 1.05 & \gradient{2.31}\( \pm \) 1.26 & \gradient{1.83}\( \pm \) 1.25 & \gradient{2.00}\( \pm \) 1.25 & \gradient{1.99}\( \pm \) 1.27 & \gradient{2.17}\( \pm \) 1.38 & \gradient{1.89}\( \pm \) 1.16 & \gradient{1.82}\( \pm \) 1.22& \\[10pt] 
			\midrule
			& FSL (GFCI) & \gradient{0.49}\( \pm \) 0.49 &\bftab \gradient{0.62}\( \pm \) 0.53 &\bftab \gradient{0.66}\( \pm \) 0.60 &\bftab \gradient{0.71}\( \pm \) 0.53 & \gradient{0.49}\( \pm \) 0.53 & \gradient{0.44}\( \pm \) 0.50 &\bftab \gradient{0.65}\( \pm \) 0.54 &\bftab \gradient{0.65}\( \pm \) 0.55 &\bftab \gradient{0.42}\( \pm \) 0.50 & \gradient{0.64}\( \pm \) 0.54 & \gradient{0.65}\( \pm \) 0.55& \\[10pt] 
			\midrule
			& Known Structure & \gradient{0.22} \( \pm \) 0.33 & \gradient{0.26} \( \pm \) 0.36 & \gradient{0.33} \( \pm \) 0.49 & \gradient{0.51} \( \pm \) 0.46 & \gradient{0.09} \( \pm \) 0.21 & \gradient{0.17} \( \pm \) 0.33 & \gradient{0.30} \( \pm \) 0.39 & \gradient{0.31} \( \pm \) 0.43 & \gradient{0.22} \( \pm \) 0.35 & \gradient{0.26} \( \pm \) 0.40 & \gradient{0.29} \( \pm \) 0.41 \\[10pt] 
		\end{tabular}
    }
    \caption{Average RMSE for different HTE estimation methods under varying experimental settings, with a dataset size of 10,000. 
    Each column shows a fixed parameter value, and averaged RMSE over all other settings.}\label{tab:rmse_table}
\end{table*}

We also show the average RMSE for each method in the same parameter settings in Table~\ref{tab:rmse_table}.
Here, Causal Tree and Balancing Neural Network achieve lower average RMSE.
DRLearner has high error and variance in predictions, which suggests it requires more fine-tuning of its estimators.

Looking at specific parameters, we see that SCMs with mediators generally result in higher errors than without mediators.
This makes sense since including mediators will bias the effect estimation.
Interestingly, a chain length of 1 is more challenging than a chain length of 2.
Increasing the number of HTE inducing parents also results in higher errors since 0 HTE parents imply average effect only, and more HTE inducing parents increases the complexity of the heterogeneous effect.
Causal Tree seems to be the least affected by parameter changes compared to other HTE estimators.

\subsection{Accuracy of decisions}

In practice, practitioners may be interested in how well an HTE estimator can assign a treatment decision to individuals, rather than the magnitude in the change in outcomes.
To investigate this, we explore how well an HTE estimator can decide whether it is beneficial to treat an individual or not.
To do this, we binarize the ground truth treatment effect into positive and negative effects.
For each HTE estimator, we binarize the predictions into positive and negative predictions.
Then, an accuracy is computed over multiple datasets, signifying how many examples are labeled correctly.

Table~\ref{tab:accuracy_table} shows the accuracy of decisions for each HTE estimator, using the same columns as in Tables~\ref{tab:mse_rankings} and~\ref{tab:rmse_table}.
In general, we see that Causal Tree achieves the highest accuracy of decisions, followed by Balancing Neural Network and FSL.
Meta-learners achieve very low accuracy when there are more invalid variables, such as when there are mediators.

\begin{table*}[t]
    \centering
    \resizebox*{\linewidth}{!}{
        \renewcommand{\minval}{0.5}
        \renewcommand{\maxval}{1.0}
        \begin{tabular}{ N P{1.9cm} | *{11}{M{1.58cm}} N }
            & &  Mediator Chain Length 0 &  Mediator Chain Length 1 &  Mediator Chain Length 2 &  Contains Confounding &  No Confounding &  0 HTE Parents &  1 HTE Parent &  2 HTE Parents &  10 Variables &  20 Variables &  30 Variables  \\
            \toprule
            & SLearner & \gradientlow{0.83} & \gradientlow{0.56} & \gradientlow{0.58} & \gradientlow{0.68} & \gradientlow{0.66} & \gradientlow{0.70} & \gradientlow{0.65} & \gradientlow{0.65} & \gradientlow{0.70} & \gradientlow{0.63} & \gradientlow{0.68}& \\[10pt] 
            & TLearner & \gradientlow{0.83} & \gradientlow{0.57} & \gradientlow{0.58} & \gradientlow{0.67} & \gradientlow{0.67} & \gradientlow{0.70} & \gradientlow{0.66} & \gradientlow{0.66} & \gradientlow{0.71} & \gradientlow{0.64} & \gradientlow{0.67}& \\[10pt] 
            & XLearner & \gradientlow{0.82} & \gradientlow{0.53} & \gradientlow{0.55} & \gradientlow{0.66} & \gradientlow{0.64} & \gradientlow{0.66} & \gradientlow{0.64} & \gradientlow{0.65} & \gradientlow{0.68} & \gradientlow{0.61} & \gradientlow{0.66}& \\[10pt] 
            & DRLearner & \gradientlow{0.82} & \gradientlow{0.58} & \gradientlow{0.59} & \gradientlow{0.67} & \gradientlow{0.67} & \gradientlow{0.71} & \gradientlow{0.65} & \gradientlow{0.66} & \gradientlow{0.71} & \gradientlow{0.64} & \gradientlow{0.67}& \\[10pt] 
            & BART & \gradientlow{0.92} & \gradientlow{0.82} & \gradientlow{0.79} & \gradientlow{0.84} & \gradientlow{0.86} & \gradientlow{0.87} & \gradientlow{0.84} & \gradientlow{0.84} & \gradientlow{0.87} & \gradientlow{0.81} & \gradientlow{0.87}& \\[10pt] 
            & Causal Tree & \gradientlow{0.94} &\bftab \gradientlow{0.93} &\bftab \gradientlow{0.89} &\bftab \gradientlow{0.86} &\bftab \gradientlow{0.97} &\bftab \gradientlow{0.95} &\bftab \gradientlow{0.91} &\bftab \gradientlow{0.91} &\bftab \gradientlow{0.96} &\bftab \gradientlow{0.92} &\bftab \gradientlow{0.90}& \\[10pt] 
            & Causal Forest & \gradientlow{0.81} & \gradientlow{0.76} & \gradientlow{0.71} & \gradientlow{0.77} & \gradientlow{0.76} & \gradientlow{0.78} & \gradientlow{0.76} & \gradientlow{0.75} & \gradientlow{0.78} & \gradientlow{0.75} & \gradientlow{0.77}& \\[10pt] 
            & Balancing Neural Network &\bftab \gradientlow{0.94} & \gradientlow{0.89} & \gradientlow{0.88} & \gradientlow{0.85} & \gradientlow{0.94} & \gradientlow{0.92} & \gradientlow{0.90} & \gradientlow{0.90} & \gradientlow{0.93} & \gradientlow{0.89} & \gradientlow{0.90}& \\[10pt] 
            & TARNet & \gradientlow{0.88} & \gradientlow{0.69} & \gradientlow{0.70} & \gradientlow{0.76} & \gradientlow{0.77} & \gradientlow{0.79} & \gradientlow{0.75} & \gradientlow{0.76} & \gradientlow{0.79} & \gradientlow{0.73} & \gradientlow{0.78}& \\[10pt] 
            \midrule
            & Known Structure & \gradientlow{0.97} & \gradientlow{0.92} & \gradientlow{0.91} & \gradientlow{0.90} & \gradientlow{0.96} & \gradientlow{0.99} & \gradientlow{0.91} & \gradientlow{0.91} & \gradientlow{0.96} & \gradientlow{0.92} & \gradientlow{0.93} \\[10pt] 
        \end{tabular}
    }
    \caption{Accuracy of decisions.}\label{tab:accuracy_table}
\end{table*}

\subsection{Comparing evaluation metrics}

Since ground truth causal effects are generally not available in real-world datasets, several heuristic HTE evaluation metrics have been developed.
We consider inverse propensity weighting (IPW)~\citep{schuler-model18}, $\tau$-risk~\citep{schuler-model18,nie-rlearner17}, plugin validation (plugin $\tau$)~\citep{saito-cv19}, and counterfactual cross-validation (CFCV)~\citep{saito-cv19}.
Each metric uses estimators for propensity scores or outcomes to compute an error.
Each metric uses the full set of features in our experiments to compute a propensity score or outcome.
We use Logistic Regression for each metric that requires an estimate of the propensity score.
For $\tau$-risk, we use Linear Regression for outcome prediction.
For plugin validation and CFCV, we use Counterfactual Regression (CFR)~\citep{shalit-icml17} for the regression function.
In order to evaluate the reliability of these heuristic metrics, we compute the Spearman rank correlation of each metric compared to the ground truth ranking.
For each dataset, we rank each learner based on the heuristic error on the test set.
Then we compute a Spearman correlation in one of two ways.
The first way computes a correlation on the entire ranking for each dataset, and an average can be computed over all datasets.
The second way computes a correlation of the rank for each learner over all the datasets.

Table~\ref{tab:metric_dataset} shows the correlation of each heuristic metric ranking to the ground truth ranking over the synthetic datasets and the semi-synthetic (IHDP) datasets.
This table shows that most metrics are positively correlated with the ranking from the ground truth on synthetic datasets, except for plugin validation.
However, there is a large variance, and the average correlation is generally low, which suggests that having a hidden structural causal model with invalid variables may affect the metrics.
For the IHDP dataset, we see that all metrics except plugin validation positively correlate with the true ranking, with $ \tau  $-risk having the highest correlation.
Since the IHDP dataset does not contain mediators or descendants of treatment, this may improve the correlation of $ \tau  $-risk.

\begin{table*}[t]
    \centering
    \resizebox*{\linewidth}{!}{
        \begin{tabular}{l *{8}{M{1.7cm}}}
            & \multicolumn{7}{c}{Synthetic} \\
            \cmidrule(lr){2-8}
            & Overall & Has Mediator & No Mediator & Contains Confounding & No Confounding & 0 HTE Parents & 1 or 2 HTE Parents &  IHDP \\
            \midrule
            IPW &  0.10 $ \pm $ 0.55 & 0.03 $\pm$ 0.59 & 0.01 $\pm$ 0.43 & 0.02 $\pm$ 0.44 & 0.08 $\pm$ 0.45 & 0.11 $\pm$ 0.42 & 0.03 $\pm$ 0.44 & 0.10 $ \pm $ 0.43 \\
            $ \tau $-risk & 0.20 $\pm$ 0.47 & 0.01 $\pm$ 0.45 & 0.10 $\pm$ 0.43 & 0.13 $\pm$ 0.44 & 0.07 $\pm$ 0.43 & 0.13 $\pm$ 0.47 & 0.09 $\pm$ 0.43 & 0.53 $ \pm $ 0.32\\
            Plugin $\tau$ & -0.18 $\pm$ 0.55 & -0.23 $\pm$ 0.56 & -0.27 $\pm$ 0.44 & -0.27 $\pm$ 0.43 & -0.25 $\pm$ 0.44 & -0.24 $\pm$ 0.45 & -0.26 $\pm$ 0.43 & -0.34 $ \pm $ 0.38\\
            CFCV & 0.25 $\pm$ 0.59 & 0.14 $\pm$ 0.61 & 0.06 $\pm$ 0.47 & 0.03 $\pm$ 0.47 & 0.03 $\pm$ 0.45 & 0.15 $\pm$ 0.56 & 0.01 $\pm$ 0.47 & 0.20 $ \pm $ 0.44 \\
        \end{tabular}
    }
    \caption{Comparing the rank correlation of metrics with true performance.}\label{tab:metric_dataset}
    \vspace{1em}
\end{table*}

\subsection{Semi-synthetic}

The IHDP dataset is a semi-synthetic dataset, where covariates are used to generate simulated outcome so that ground truth effects are available for evaluation.
We show two results in Table~\ref{tab:ihdp_rankings}.
The first column of Table~\ref{tab:ihdp_rankings} shows the average rankings on the IHDP dataset in terms of RMSE.
From this Table, we see that FSL, TLearner, and XLearner perform consistently well in this dataset in terms of RMSE compared to all other estimator, since their rank is much higher than the next best performing estimator, SLearner.
The second column shows the $ \tau  $-risk results.
We see that the $ \tau  $-risk rank is generally correlated with the RMSE rank of each learner.
A notable difference is that Causal Forest is ranked higher based on $ \tau  $-risk, but lower based on RMSE.

\begin{table}[!t]
    \centering
    \resizebox*{1.0\linewidth}{!}{
		\begin{tabular}{ N P{2.25cm} | *{5}{M{2.0cm}} N }
			& & IHDP MSE Rank &  IHDP $\tau$-risk Rank  & LoL $\tau$-risk Rank & Cannabis $\tau$-risk Rank & SHARPS $\tau$-risk Rank \\
			\midrule
			& SLearner & 5.22 \( \pm \) 1.34 & 6.42 \( \pm \) 1.82 & 7.85 $ \pm $ 0.70 & 7.17 $ \pm $ 0.83 & \bftab 1.29 $ \pm $ 0.48 & 
			\\[10pt]
			& TLearner & 2.89 \( \pm \) 1.41 & 2.68 \( \pm \) 1.52 & 2.95 $ \pm $ 0.59 & 2.74 $ \pm $ 0.82 & 6.59 $ \pm $ 0.50 & 
			\\[10pt]
			& XLearner & 3.07 \( \pm \) 1.18 & 3.03 \( \pm \) 1.40 & 2.08 $ \pm $ 0.51 & \bftab 1.51 $ \pm $ 0.63 & 6.57 $ \pm $ 0.50 & 
			\\[10pt]
			& DRLearner & 6.12 \( \pm \) 2.33 & 6.41 \( \pm \) 1.93 & 5.56 $ \pm $ 0.50 & 7.62 $ \pm $ 0.60 & 8.08 $ \pm $ 0.00 & 
			\\[10pt]
			& BART & 5.92 \( \pm \) 1.72 & 5.63 \( \pm \) 2.06 & 4.89 $ \pm $ 0.45 & 2.74 $ \pm $ 0.50 & 5.41 $ \pm $ 0.76 &
			\\[10pt]
			& Causal Tree & 6.21 \( \pm \) 1.93 & 6.00 \( \pm \) 2.18 & 4.80 $ \pm $ 0.36 & 2.14 $ \pm $ 0.80 & 4.58 $ \pm $ 0.82 & 
			\\[10pt]
			& Causal Forest & 6.39 \( \pm \) 1.72 & 4.00 \( \pm \) 2.06 & \bftab 1.81 $ \pm $ 0.45 & 3.94 $ \pm $ 0.50 & 4.30 $ \pm $ 0.76 & 
			\\[10pt]
			& Balancing Neural Network & 6.55 \( \pm \) 1.34 & 6.82 \( \pm \) 1.91 & 3.17 $ \pm $ 0.54 & 6.17 $ \pm $ 0.77 & 2.69 $ \pm $ 0.69 & 
			\\[10pt]
			& TARNet & 7.17 \( \pm \) 1.38 & 5.98 \( \pm \) 1.97 & 3.53 $ \pm $ 0.51 & 5.32 $ \pm $ 0.59 & 2.53 $ \pm $ 0.83 & \\[10pt]
			& FSL (O-set) & \bftab 1.97 \( \pm \) 1.94 & \bftab 2.07 \( \pm \) 2.34 & 4.53 $ \pm $ 0.47 & 3.90 $ \pm $ 0.61 & 3.53 $ \pm $ 0.83 & 
			\\[10pt]
		\end{tabular}	
    }
    \caption{Average rank of estimation methods for the IHDP, LoL, and Cannabis datasets.}\label{tab:ihdp_rankings}
    \vspace{-1em}
\end{table}

\subsection{Real-world datasets}

In real-world datasets, ground truth effects are not available.
Instead, we opt to compare HTE estimators based on $ \tau  $-risk as a proxy for methods perform in different datasets.
The average rank for each method in terms of $\tau$-risk for the LoL dataset is shown in Table~\ref{tab:ihdp_rankings}.
In this dataset, Causal Forest outranks other learners by a significant margin, followed by XLearner.
The rank in terms of $\tau$-risk for the Cannabis dataset is shown in Table~\ref{tab:ihdp_rankings}.
In the Cannabis dataset, XLearner achieves the highest $ \tau $-risk rank consistently. 
Finally, the $\tau$-risk for the SHARPS dataset is shown in the last column of Table~\ref{tab:ihdp_rankings}.
In the SHARPS dataset, SLearner achieves the best rank consistently.
An interesting observation from this dataset is that the average difference in outcomes between treated and non-treated individuals is a small number (close to zero).
As pointed out by K{\"{u}}nzel et al., SLearner has a tendency to shrink estimates toward zero~\citep{kunzel-pnas19}, which may play a role in the $\tau$-risk rank in this dataset.

\section{Summary and discussion}\label{sec:conclusion}

In this paper, we provide an introduction to data-driven heterogeneous treatment effect estimation and an overview of data-driven estimators that have been developed.
We broadly categorized data-driven estimators into counterfactual prediction methods and effect estimation methods.
To explore how these data-driven methods perform against each other, we explored their use in synthetic, semi-synthetic, and real-world datasets.
In general, we found that the performance of HTE estimators varied depending on the datasets, and there is no straightforward optimal method to use in all scenarios. 

A current limitation of data-driven HTE estimators is the assumption of ignorability.
Our experiments show that without knowledge of the underlying model, the error of HTE estimations will increase, as shown in synthetic experiments.
This weakness can be more pronounced in real-world datasets, and addressing how to use the underlying model in HTE estimation is an essential avenue for future research.
Besides the assumption of ignorability, selection bias~\citep{heckman-jes79,cortes-alt08,bareinboim-pnas16} in the data can impact the accuracy of estimation. Some work has been done to address accounting for selection bias in causal models~\citep{bareinboim-pnas16}, which further emphasizes the importance of using causal models with data-driven estimation.
In addition, understanding the causal model, evaluation of current models currently has a large gap due to a lack of benchmark datasets.
A recent paper points out this gap and suggests alternatives for metrics and benchmark datasets~\citep{curth-neurips21} and more research in this area is an important future direction. 
Finally, as we have explored in our experiments, heuristic evaluation metrics are not always reliable.
Understanding when these metrics can be used and how to improve them is another fruitful avenue of research.

\bibliographystyle{plain}
\bibliography{references}

\section*{Disclaimer}
\textit{The Securities and Exchange Commission disclaims responsibility for any private publication or statement of any SEC employee or Commissioner. This article expresses the author's views and does not necessarily reflect those of the Commission, the Commissioners, or members of the staff. This paper was initially released prior to the author joining the Commission}

\end{document}